\documentclass[authoryear,10pt]{article}
\usepackage[margin=2cm]{geometry}
\usepackage{cite}
\usepackage{amsmath,amssymb,amsfonts}
\usepackage{algorithmic}
\usepackage{graphicx}
\usepackage{textcomp}
\usepackage{float}
\usepackage{natbib}
\usepackage{color}
\usepackage{xcolor}
\usepackage{hyperref}
\usepackage[T1]{fontenc}
\usepackage{tgheros}
\usepackage{parskip}
\usepackage{caption}
\usepackage{subcaption}
\usepackage{tikz}
\usetikzlibrary{shapes,arrows,positioning,calc}
\usepackage[capitalise]{cleveref}
\usepackage{booktabs}
\usepackage{multirow}
\usepackage{mathtools}
\usepackage{setspace}
\usepackage[T1]{fontenc}

\hypersetup{
    colorlinks,
    linkcolor={blue!50!black},
    citecolor={blue!50!black},
    urlcolor={blue!50!black}
}

\title{Robust image representations\\with counterfactual contrastive learning}

\author{Mélanie Roschewitz$^1$, Fabio De Sousa Ribeiro$^1$, Tian Xia$^1$, Galvin Khara$^2$, Ben Glocker$^{1,2}$\\
\small{$^1$Imperial College London, $^2$Kheiron Medical Technologies}}
\date{}

\begin{document}

\maketitle

\begin{abstract}
Contrastive pretraining can substantially increase model generalisation and downstream performance. However, the quality of the learned representations is highly dependent on the data augmentation strategy applied to generate positive pairs. Positive contrastive pairs should preserve semantic meaning while discarding unwanted variations related to the data acquisition domain. Traditional contrastive pipelines attempt to simulate domain shifts through pre-defined generic image transformations. However, these do not always mimic realistic and relevant domain variations for medical imaging, such as scanner differences. To tackle this issue, we herein introduce \emph{counterfactual contrastive learning}, a novel framework leveraging recent advances in causal image synthesis to create contrastive positive pairs that faithfully capture relevant domain variations. Our method, evaluated across five datasets encompassing both chest radiography and mammography data, for two established contrastive objectives (SimCLR and DINO-v2), outperforms standard contrastive learning in terms of robustness to acquisition shift. Notably, counterfactual contrastive learning achieves superior downstream performance on both in-distribution and external datasets, especially for images acquired with scanners under-represented in the training set. Further experiments show that the proposed framework extends beyond acquisition shifts, with models trained with counterfactual contrastive learning reducing subgroup disparities across biological sex.
\end{abstract}

\section{Introduction}
Contrastive learning in medical imaging has emerged as an effective strategy to leverage unlabelled data. This self-supervised learning approach has been shown to substantially improve model generalisation across domain shifts as well as reduce the amount of high-quality annotated data needed for training~\citep{azizi_big_2021,azizi_robust_2023,ghesu_contrastive_2022,zhou_foundation_2023}. However, the success of contrastive-based learning is heavily dependent on the positive pair generation pipeline~\citep{tian_what_2020}. These positive pairs are typically generated by repeatedly applying pre-defined data augmentations to the original image. As such, changes in the augmentation pipeline have a substantial impact on the quality of the learned representations, ultimately influencing downstream performance and robustness to domain changes~\citep{tian_what_2020,scalbert_towards_2023}. Traditionally, augmentation pipelines developed for natural images have been directly applied to medical imaging, however, this might not be optimal due to the unique challenges and characteristics of how medical scans are acquired. In particular, domain variations are often much larger than subtle class-wise differences. This may lead contrastively-learned representations to inadvertently encode these irrelevant acquisition-related variations into the learned representations.

\begin{figure}[t]
    \centering
    \includegraphics[width=\textwidth]{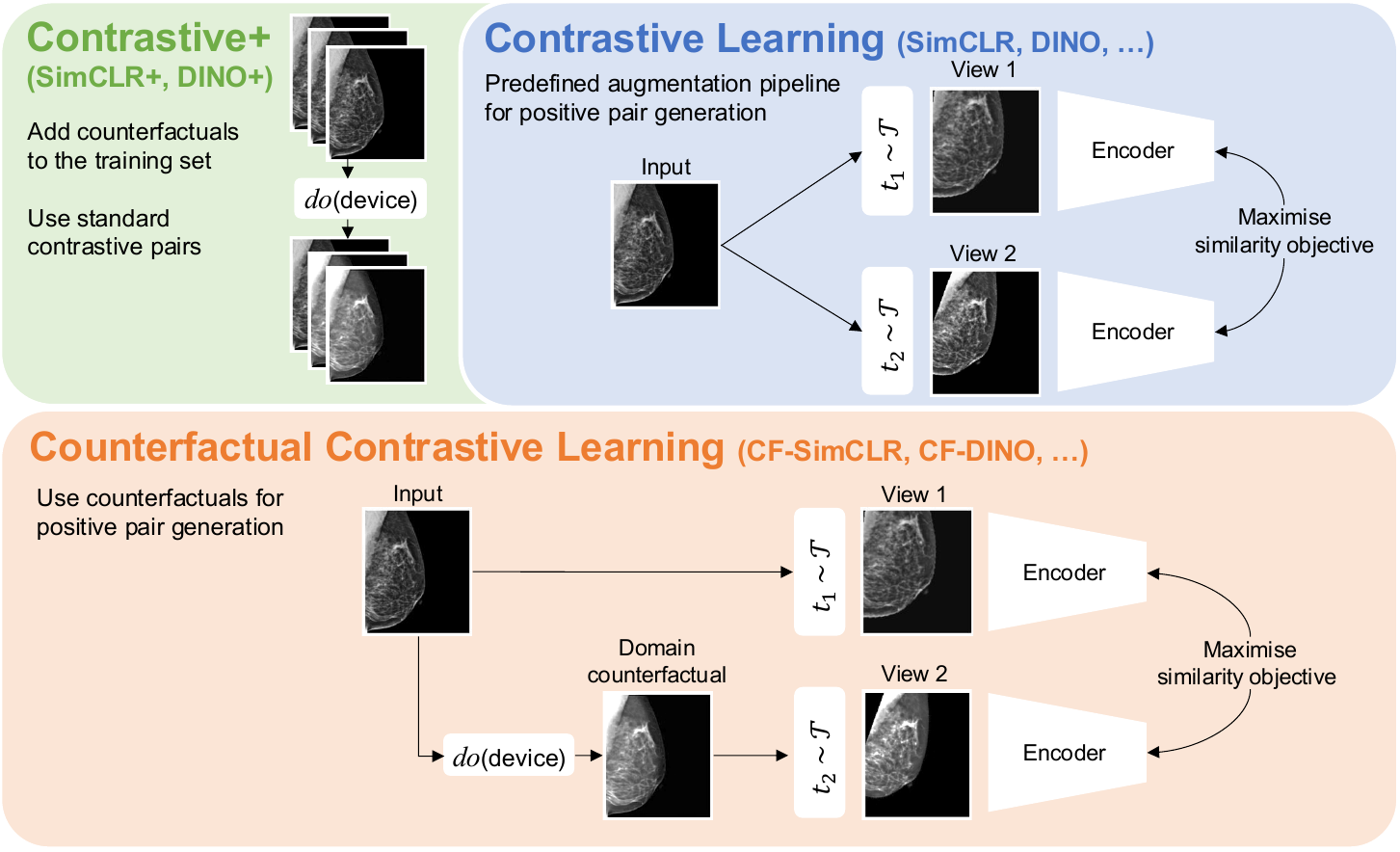}
    \caption{\textbf{We propose a novel counterfactual contrastive pair generation framework for improving robustness of contrastively-learned features to distribution shift}. As opposed to solely relying on a pre-defined augmentation generation pipeline $\mathcal{T}$ (as in standard contrastive learning), we propose to combine real images with their domain counterfactuals to create realistic cross-domain positive pairs. Importantly, this proposed approach is independent of the specific contrastive objective employed. The causal image generation model is represented by the `do’ operator. We also compare the proposed method to another approach where we simply extend the training set with the generated counterfactual images without explicit matching with their real counterparts, treating real and counterfactuals as independent training samples. Figure adapted from~\citep{conferencepaper}.}
    \label{fig:fig1}
\end{figure}

In this work, we aim to improve the robustness of contrastively-learned representation against domain shifts, in particular acquisition shift. Acquisition shift is caused by changes in image acquisition protocols (device settings, post-processing software, etc.), and is a major source of dataset shift in the medical imaging domain. 
We hypothesise that the robustness of contrastively-learned features against such changes in image characteristics could be improved by simulating domain variations more faithfully in the positive pair creation stage. For this reason, we propose and evaluate `counterfactual contrastive learning’, a new contrastive pair generation framework leveraging recent advances in deep generative models for high-quality, realistic counterfactual image generation~\citep{ribeiro_high_2023,fontanella_diffusion_2023}. Counterfactual generation models allow us to answer `what-if' questions, such as simulating how a mammogram acquired with one device would appear if it had been acquired with a different device. Specifically, in our proposed counterfactual contrastive framework, we create cross-domain positive pairs, by matching real images with their domain counterfactuals, realistically simulating device changes. Importantly, the proposed approach is agnostic to the choice of the contrastive objective as it only impacts the positive pair creation step. We illustrate the benefits of this approach for two widely-used contrastive learning frameworks: seminal work SimCLR~\citep{chen_simple_2020} and newly released DINO-V2~\citep{oquab_dinov2_2023} objectives. Moreover, to precisely measure the effect of the proposed counterfactual pair generation process, we also compare the proposed approach to a simpler approach where we simply extend the training set with the generated counterfactuals.

Evaluating the proposed counterfactual contrastive learning framework across two medical image modalities, mammography and chest radiographs, on five public datasets and two clinically-relevant classification tasks, we show that our method yields features which are more robust to domain changes. This increased robustness was directly observed in the feature space and, more importantly, by a substantial increase in downstream performance, in particular under limited labels and for domains under-represented at training time. Crucially, these findings hold for both SimCLR and DINO-V2, despite major differences in the training objectives.
This paper is an extension of our recent MICCAI workshop paper~\citep{conferencepaper}. It differs in the following aspects:
\begin{itemize}
    \item we previously only considered the SimCLR objective, here we extend our counterfactual contrastive approach to the recently proposed DINO-V2~\citep{oquab_dinov2_2023} objective. These new results demonstrate empirically that the proposed approach is versatile and agnostic to the choice of the contrastive objective.
    \item While the main focus of this work is robustness to acquisition shift, in this extension, we show that the proposed method can be extended beyond this scenario, for example, to improve subgroup performance.
    \item We substantially expanded the discussion, methods and related work sections. 
\end{itemize}

\section{Related work}

\subsection{Contrastive learning} 
\label{sec:cl}
Generating pairs of image `views' that share the same underlying meaning (positive pairs) is the core principle of contrastive learning. The contrastive objective then encourages the model to learn similar embeddings for these pairs, while keeping them distinct from the embeddings of unrelated images. A landmark work in this field is SimCLR~\citep{chen_simple_2020}, where positive pairs are generated by applying photometric and geometric transformations to the original image. SimCLR stands out for its simplicity, effectiveness, and widespread adoption, particularly in medical imaging. \citet{azizi_robust_2023} for example showed that pre-training models with SimCLR substantially improves downstream performance and robustness to various sources of data shifts. Several methods have proposed refinements to SimCLR, for example BYOL~\citep{grill_bootstrap_2020}, MoCo~\citep{he_momentum_2020}, or most recently DINO(-v2)~\citep{caron_emerging_2021,oquab_dinov2_2023}. DINO uses a self-distillation approach without explicit negative pairs, where a student network learns to match the output of a teacher network, updated via momentum. This method focuses on consistency between the teacher and student outputs, making it less dependent on batch size and augmentations. Importantly positive pairs are generated using more than two views. Representations are instead encouraged to be similar across different global views (larger image crops) and local views (smaller crops). Moreover, DINO relies on vision transformers~\citep{dosovitskiy_image_2020}, contrarily to SimCLR which was primarily designed for convolutional networks. Further enhancements were proposed in DINO-v2~\citep{oquab_dinov2_2023}, with modifications to the loss function to improve stability and performance, encouraging better feature alignment and consistency. Recent work has successfully applied DINO-v2 pre-training to chest radiography, achieving state-of-the-art downstream performance across different tasks~\citep{moutakanni_advancing_2024}.

Contrastive learning is a popular paradigm in self-supervised learning for medical imaging, in particular SimCLR which was found to be the most used contrastive objective in a recent review by~\citet{huang_self-supervised_2023}. While most works use standard positive pair creation (designed for natural images)~\citep{azizi_big_2021,azizi_robust_2023,moutakanni_advancing_2024}, some have derived medical imaging-specific positive pair creation methods such as using neighbouring slices in CT images~\citep{dong_case_2021}, neighbouring patches in histopathology~\citep{li_dual-stream_2021}, leveraging multiple views in mammography~\citep{linguraru_mammoclip_2024}, or using patient metadata to find similar images~\citep{vu_medaug_2021,dufumier_contrastive_2021}. Another approach consists of leveraging multi-modal patient information to form positive pairs, e.g. reports and chest X-rays~\citep{zhang_contrastive_2022,linguraru_mammoclip_2024} or combining imaging with genetic information~\citep{taleb_contig_2021}. However, such paired views or multi-modal pairs may not always be available, further motivating our investigation of the use of synthetic pairs for medical image contrastive learning.

Previously, \citet{von_kugelgen_self-supervised_2021} have theoretically demonstrated that contrastive learning provably disentangles content from style, provided that augmentations capture realistic style changes, proposing to interpret data augmentation in contrastive settings as `counterfactuals under soft style intervention' and providing a strong theoretical motivation to our work.

\subsection{Counterfactual image generation}

One goal of counterfactual image generation is to produce `counterfactual explanations', i.e. images depicting the smallest change in the input that would have changed the prediction of a pre-defined classifier~\citep{augustin_diffusion_2022,sanchez_what_2022,atad_chexplaining_2022,matsui_counterfactual_2022,sun_inherently_2024}. Parallel to this interpretability-centered line of work, others have focused on using generative modelling to synthesise `what-if' images, independently of any external classifier. Seminal work by~\citet{pawlowski_deep_2020} introduced Deep Structural Causal Models (DSCM) to generate realistic counterfactuals for small-resolution images. This framework has been substantially extended by~\citet{ribeiro_high_2023}, where the authors utilise a hierarchical variational autoencoder (HVAE) for improving image generation, unlocking high-quality high-resolution counterfactual generation, in particular for medical images. While counterfactual image generation models are gaining traction, with some studies showing promising results in imbalanced data augmentation~\citep{xia_adversarial_2022,garrucho_high-resolution_2023} and fairness~\citep{dash_evaluating_2022}, the potential of these models for enhancing performance on clinically relevant tasks still warrants more exploration.

\subsection{Combining contrastive learning and counterfactuals} 
\citet{zhang_counterfactual_2020} explored the use of counterfactual text-image pairs in vision-language grounding tasks, defining task-dependent counterfactual pairs for additional supervision signal. Within the context of supervised contrastive graph learning, \citet{yang_generating_2023} proposed to generate challenging negative examples using graph counterfactuals. However, the use of image-only counterfactuals for vision contrastive learning remains largely unexplored.

\section{Counterfactual contrastive learning}
In this section, we introduce \emph{counterfactual contrastive learning}, a self supervised learning paradigm to train image encoders robust to domain variations by leveraging state-of-the-art counterfactual image generation models. Contrastive learning typically uses colour and intensity-based image augmentations to encourage the model to ignore domain-specific image characteristics. However, in medical imaging, the effect of changes in acquisition hardware, device calibration or post-processing software on the final image appearance is highly complex and can not realistically be replicated by those simple handcrafted transformations. To overcome this, in counterfactual contrastive learning, we instead use a causal image generation model to simulate realistic domain variations and generate cross-domain contrastive pairs, explicitly encouraging contrastively-learned representations to ignore domain-specific information (such as scanner differences). We illustrate key differences between standard contrastive learning and the proposed counterfactual contrastive approach in \cref{fig:fig1}.

\subsection{Counterfactual image generation model}
Formally, a Structural Causal Model~\citep{pearl_causality_2009} (SCM) is defined by a triple $\mathcal{M} = \langle \mathbf{X}, \mathbf{U}, \mathbf{F} \rangle$, where $\mathbf{X} =\{X_{i}\}_{i=1}^n$ represents the set of endogenous (observed) variables, $\mathbf{U} = \{U_{i}\}_{i=1}^n$ a set of exogenous (unobserved) variables, and $\mathbf{F} = \{f_{i}\}_{i=1}^n$ a set of functions, aka causal mechanisms, such that $X_{k}:=f_{k}(\mathbf{pa}_{k}, U_{k})$, where $\mathbf{pa}_{k}\subseteq \mathbf{X} \setminus X_{k}$ are called parents (i.e. direct causes) of each $X_{k}$. The variables in $\mathbf{X}$ are called endogenous since they are caused by the variables in the model, whereas variables in $\mathbf{U}$ are exogenous as they are caused by factors which are external to the model. The process of generating counterfactuals can then be divided into three steps:
\begin{itemize}
    \item Abduction: infer the posterior exogenous noise distribution $P_{\mathbf{U}\mid\mathbf{X}}$ given observed evidence $\mathbf{X}$;
    \item Intervention: perform an intervention by modifying one or more of the endogenous variables, e.g. $do(X_{k}:= x)$, to obtain a modified SCM $\mathcal{M}_x$;
    \item Prediction: use $\mathcal{M}_x$ and $P_{\mathbf{U}\mid\mathbf{X}}$ to compute a counterfactual.
\end{itemize}

Here, we use the Deep Structural Causal Model (DSCM) proposed by~\citet{ribeiro_high_2023} to generate image counterfactuals. In what follows, we detail the counterfactual image generation process using this model. For simplicity, we here assume no further causal relationships between the parents of the image (i.e. all image parents are independent of each other). Formally, let $\mathbf{x}$ be the image and $ \mathbf{pa_{x}}$ the parents of $\mathbf{x}$. In~\citet{ribeiro_high_2023}, the mechanism $\mathbf{x} \coloneqq f_{\mathbf{x}}(\mathbf{pa}_{\mathbf{x}}, \mathbf{u}_{\mathbf{x}})$ is modelled using an HVAE. Specifically, the exogenous noise is decomposed into two parts $p(\mathbf{u}_{\mathbf{x}}) = p_\theta(\mathbf{z})p(\boldsymbol{\epsilon})$, where the first is inferred using the HVAE's encoder $q_{\phi}(\mathbf{z} \mid \mathbf{x}, \mathbf{pa}_{\mathbf{x}})$, and the second by inverting the decoder $g_\theta$'s sampling mechanism $\boldsymbol{\epsilon} = (\mathbf{x} - \boldsymbol{\mu}(\mathbf{z}, \mathbf{pa}_{\mathbf{x}})) / \boldsymbol{\sigma}(\mathbf{z}, \mathbf{pa}_{\mathbf{x}})$, where $\boldsymbol{\mu}$ and $\boldsymbol{\sigma}$ here are per pixel mean/std predictions by the decoder. To generate a counterfactual image $\widetilde{\mathbf{x}}$, we simply compute $\widetilde{\mathbf{x}} = \boldsymbol{\mu}(\mathbf{z}, \widetilde{\mathbf{pa}}_{\mathbf{x}}) + \boldsymbol{\sigma}(\mathbf{z}, \widetilde{\mathbf{pa}}_{\mathbf{x}}) \odot \boldsymbol{\epsilon}$ holding the exogenous noise fixed, where $\widetilde{\mathbf{pa}}_{\mathbf{x}}$ are the counterfactual parents obtained after modifying one or more of $\mathbf{pa}_{\mathbf{x}}$'s values. Note that in this work, we use the conditional prior proposed by~\citet{ribeiro_high_2023} in the HVAE, i.e.
 $p_{\theta}(\mathbf{z}_{1:L} \mid \mathbf{pa}_{\mathbf{x}}) = p_{\theta}(\mathbf{z}_{L} \mid \mathbf{pa}_{\mathbf{x}})\prod_{i=1}^{L-1}p_{\theta}(\mathbf{z}_{i} \mid \mathbf{z}_{>i},\mathbf{pa}_{\mathbf{x}}).$ The HVAE is then trained by maximising the Evidence Lower Bound (ELBO) of the log-likelihood on the observed dataset. In this work, minor modifications were made to the original HVAE model from~\citet{ribeiro_high_2023} to further increase training stability and image quality. First, instead of directly using the parent variables to condition the HVAE, we add an embedding layer to learn a more flexible parents' embedding for improved conditioning. Moreover, we used SiLU~\citep{hendrycks_gaussian_2023} activation layers instead of ReLU, added group normalisation layers, and fixed the decoder's variance to 1e-2, as we noticed that these changes improved training stability.

Note that, in their initial study,~\citet{ribeiro_high_2023} noticed that solely relying on likelihood training for the HVAE may sometimes lead to `ignored counterfactual conditioning' i.e. $\mathbf{\widetilde{x}}$ does not obey $\mathbf{\widetilde{pa}_{x}}$. To mitigate this issue, the authors introduced counterfactual finetuning. This step is optional and involves further finetuning the HVAE weights after the likelihood training in order to improve \emph{effectiveness} of the model (i.e. how well the generated counterfactual obeys the counterfactual parents). Concretely, this method leverages a pre-trained classifier $q_{\psi}(\mathbf{\widetilde{pa}_{x}} \mid \widetilde{\mathbf{x}})$, and optimises the pre-trained HVAE weights $\{\theta,\phi\}$ to maximise $\log q_{\psi}(\mathbf{\widetilde{pa}_{x}} \mid \widetilde{\mathbf{x}})$ with $\psi$ fixed. However,~\citet{xia_mitigating_2024} showed that this finetuning step may overly emphasise non-intervened attributes, a phenomenon known as `attribute amplification', and it can be mitigated by using \textit{soft} rather than \textit{hard} labels during finetuning. In this study, we found that our generation model achieved sufficient image quality without relying on any counterfactual finetuning. Hence, unless otherwise stated, the counterfactual image generation used in the following does not rely on this step.

\subsection{A simple and effective approach to counterfactual contrastive learning: CF-SimCLR}
\label{sec:cfsimclr}
SimCLR~\citep{chen_simple_2020} is a widely adopted contrastive learning strategy, due to its effectiveness and simplicity in terms of training setup and number of hyperparameters to tune. Contrastive pairs are composed of two related views from the same image, obtained by applying a random augmentation pipeline to the original input. An image encoder then yields a high dimensional representation for each of the views, which is then projected onto a lower dimensional space using a two-layer perceptron to obtain representations $z$. The NT-Xent loss then pushes the representations of positive pairs closer together, while representations of negative pairs are pushed apart. Concretely, for each positive pair $(i,j)$, the loss is given by:
\begin{align}
 \mathcal{L}_{i,j} = - \log \frac{\exp(\text{sim}(z_i, z_j)/\tau)}{\sum_{k=1, k\ne i}^{2N} \exp(\text{sim}(z_i, z_k)/\tau)}, && \text{sim}(u,v) = \frac{u^{T}v}{\|u\|\|v\|}.   
\end{align}

In this work, we propose a novel approach to contrastive positive pair creation, relying on domain counterfactuals instead of pre-defined random image transformations only. Applying this novel counterfactual pair creation pipeline to the classic SimCLR objective, we obtain `CF-SimCLR', where we create positive view pairs by pairing each real image with one of its domain counterfactuals. Concretely, we sample one target domain at random among all possible domains and generate the corresponding domain counterfactual to pair with the real image. If the original domain is sampled, we simply keep the real image as the domain counterfactual (since there are no domain changes). To further increase view diversity, we then apply the original augmentation pipeline to this cross-domain positive pair. The rest of the SimCLR framework is kept as-is, as summarised in \cref{fig:fig1}.

The proposed method uses synthetically generated images for model training, in addition to the real images. Hence, the effective training size increases significantly compared to a model trained on real data only. As such, we need to disentangle the effect of the `smart pair creation' proposed in CF-SimCLR and the effect of synthetically increasing the training set size. For this purpose, we here introduce another baseline, SimCLR\texttt{+}, where we add the same amount of synthetic examples in the training set as in CF-SimCLR. However, in this baseline, we do not implement our counterfactual pairing strategy: all images are considered independent samples, and standard contrastive learning is applied. As such, SimCLR\texttt{+} and CF-SimCLR are trained using the exact same training set; only the positive pair creation mechanism differs. SimCLR\texttt{+} is illustrated in green in \cref{fig:fig1}.

\subsection{Extension to other contrastive objectives: CF-DINO}
\label{CF-DINO}
The proposed counterfactual contrastive framework defines a novel way to create contrastive pairs by leveraging counterfactual image generation, independently of the particular choice of contrastive objective. To demonstrate that our counterfactual contrastive framework is indeed general and directly applicable to other contrastive learning objectives, we here extend our counterfactual contrastive analysis to models trained with the recently proposed and popular DINO-v2 objective~\citep{oquab_dinov2_2023}. As introduced in \cref{sec:cl}, DINO-v2 combines contrastive losses over different crops of images, model distillation and vision transformers to learn general image representations. For each image, we generate two `global views' (i.e. larger crops of the images) as well as eight additional `local views' (i.e. smaller crops). The model is then encouraged to produce similar representations for all global and local crops for both the student and teacher models. The final loss function is also complemented by a masked image modelling component. We invite the reader to refer to the original DINO-v2 paper for further details~\citep{oquab_dinov2_2023}.

To incorporate our counterfactual contrastive strategy with DINO-v2, we follow steps similar to those of CF-SimCLR. Specifically, in `CF-DINO', one global crop is created from the real image while the other one is generated from its counterfactual. Similarly, we generate half of the local crops from the real image (N=4) and half from its counterfactual (N=4). During training, all views are encouraged to produce similar image representations, yielding the desired cross-domain invariance.

\section{Experiments}
\subsection{Datasets} 
We evaluate the proposed method on two medical image modalities, mammography and chest radiography, using five public datasets covering a large variety of image acquisition hardware. The main chest radiography dataset used in this study is PadChest~\citep{bustos_padchest_2020}, a large dataset from Spain composed of scans acquired with two different scanners. In this dataset, scanner information is available for every image, allowing us to train a domain counterfactual generation model easily. We use the same dataset for self-supervised pretraining. We evaluate the quality of the learned representations on pneumonia detection, first on in-distribution PadChest test data, then on two external datasets (covering acquisition domains unseen during pretraining): RSNA Pneumonia Detection~\citep{stein_rsna_2018,shih_augmenting_2019} and CheXpert~\citep{irvin_chexpert_2019}. For mammography, we primarily use the EMBED~\citep{jeong_emory_2023} dataset, containing over 300k scans, acquired in the US with 6 different devices. We keep one domain as a hold-out domain (`Senographe Essential') and use the remaining five scanners for pretraining and counterfactual image generation. We highlight that 90\% of the EMBED data was acquired with the `Selenia Dimensions' scanner, the other scanners being heavily under-represented in this dataset, an ideal setup for investigating robustness to domain shifts. Finally, we investigate the quality of the learned encoders when transferring to the external VinDR-Mammo~\citep{nguyen_vindr-mammo_2023} dataset from Vietnam, covering two different acquisition domains. \cref{tab:dataset_splits} details dataset splits and inclusion criteria.

\begin{table}
    \centering
    \begin{tabular}{llccc}
        \toprule
         \multirow{2}{*}{Dataset}  & \multirow{2}{*}{Inclusion criteria} & \multicolumn{3}{c}{Number of images} \\
         & & Train & Validation & Test \\
         \toprule
        EMBED & 2D only$^{(*)}$ &   223,086 &  8,295 & 65,992 \\
        Senographe Essential & - &  10,927 &  1,251 & 3,022 \\
        VinDR Mammo & - & 11,191 & 2,813 & 5,996 \\
        PadChest & Adult PA only & 64,989 & 7,203 & 17,993 \\
        CheXpert & PA only & 13,811 & 2,449 & 10,838 \\
        RSNA Pneumonia & PA only & 8,633 & 1,524 & 4,354 \\
        \bottomrule
    \end{tabular}
    \caption{\textbf{Datasets splits and inclusion criteria}. Splits are created at the patient level. $^{(*)}$ excluding Senographe Essential, kept as a separate hold-out domain.}
    \label{tab:dataset_splits}
\end{table}

\subsection{Causal inference model and synthetic data generation}
\label{sec:cf_gen}

\begin{figure}
\centering
\begin{subfigure}[t]{0.45\textwidth}
\centering
\begin{tikzpicture}[
    node distance=2cm,
    roundnode/.style={circle, draw=black, thick, minimum size=15mm},
    arrow/.style={->, >=stealth, thick}
]
\small
    \node[roundnode] (scanner) {Scanner};
    \node[roundnode] (x) [below =8mm of scanner] {Image};
    
    \draw[arrow] (scanner) -- (x);
\end{tikzpicture}
\caption{EMBED}
\end{subfigure}
\begin{subfigure}[t]{0.45\textwidth}
\centering

\begin{tikzpicture}[
    node distance=2cm,
    roundnode/.style={circle, draw=black, thick, minimum size=15mm},
    arrow/.style={->, >=stealth, thick}
]
\small
    \node[roundnode] (scanner) {Scanner};
    \node[roundnode] (sex) [right =12mm of scanner] {Sex};
    
    \node[roundnode] (x) [below=16mm of $(scanner)!0.5!(sex)$] {Image};
    
    \draw[arrow] (scanner) -- (x);
    \draw[arrow] (sex) -- (x);
\end{tikzpicture}
\caption{PadChest}
\end{subfigure}
\caption{\textbf{Causal graphs used to train the counterfactual image generation models used in this study.}}
\label{fig:causalgraphs}
\end{figure}

\begin{figure}
  \centering
  \hfill
  \begin{subfigure}[t]{0.6\textwidth}
  \centering
    \includegraphics[height=7.5cm]{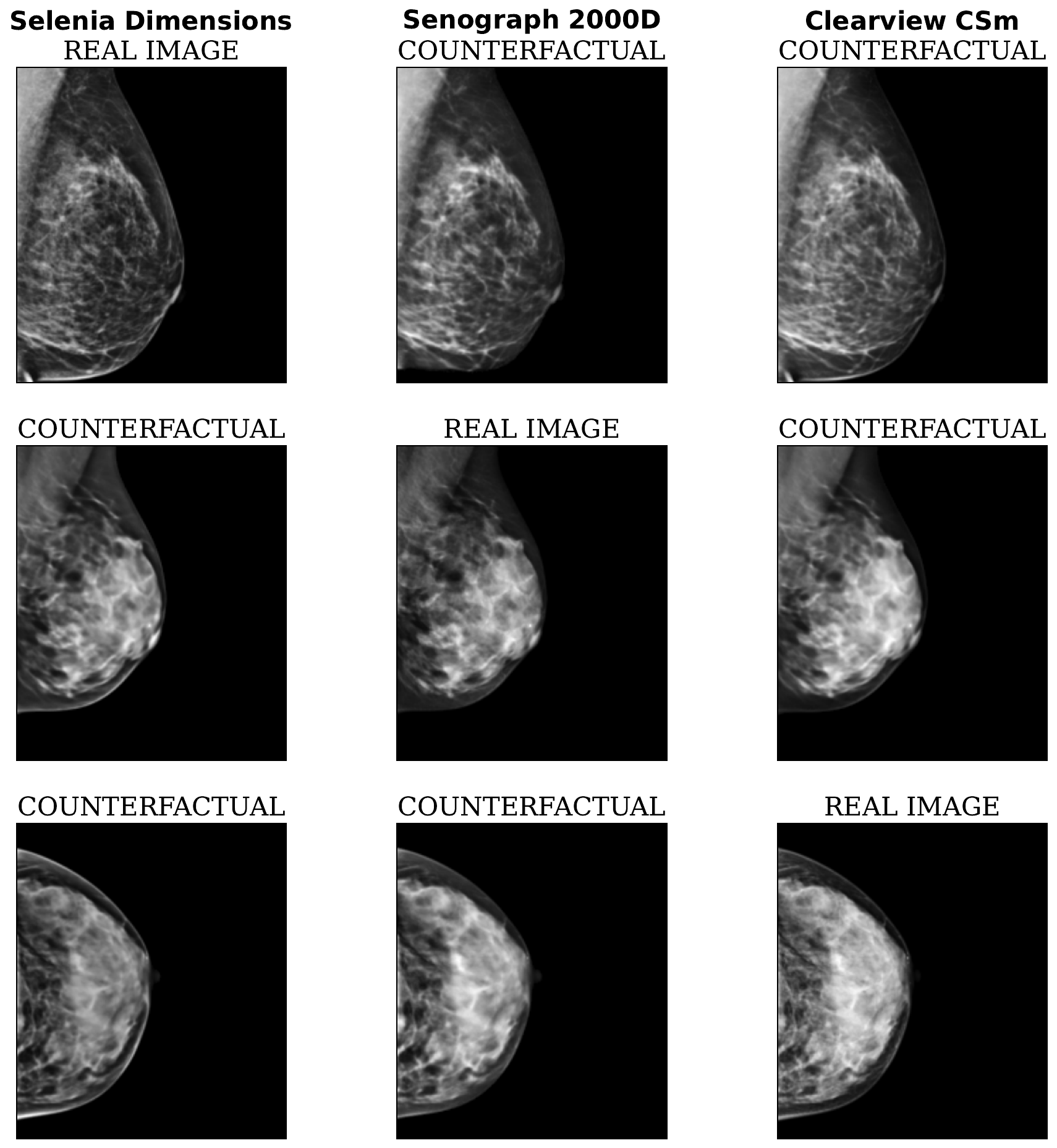}
    \caption{EMBED}
  \end{subfigure}
  \hfill
  \begin{subfigure}[t]{0.39\textwidth}
  \centering
    \includegraphics[height=7.5cm]{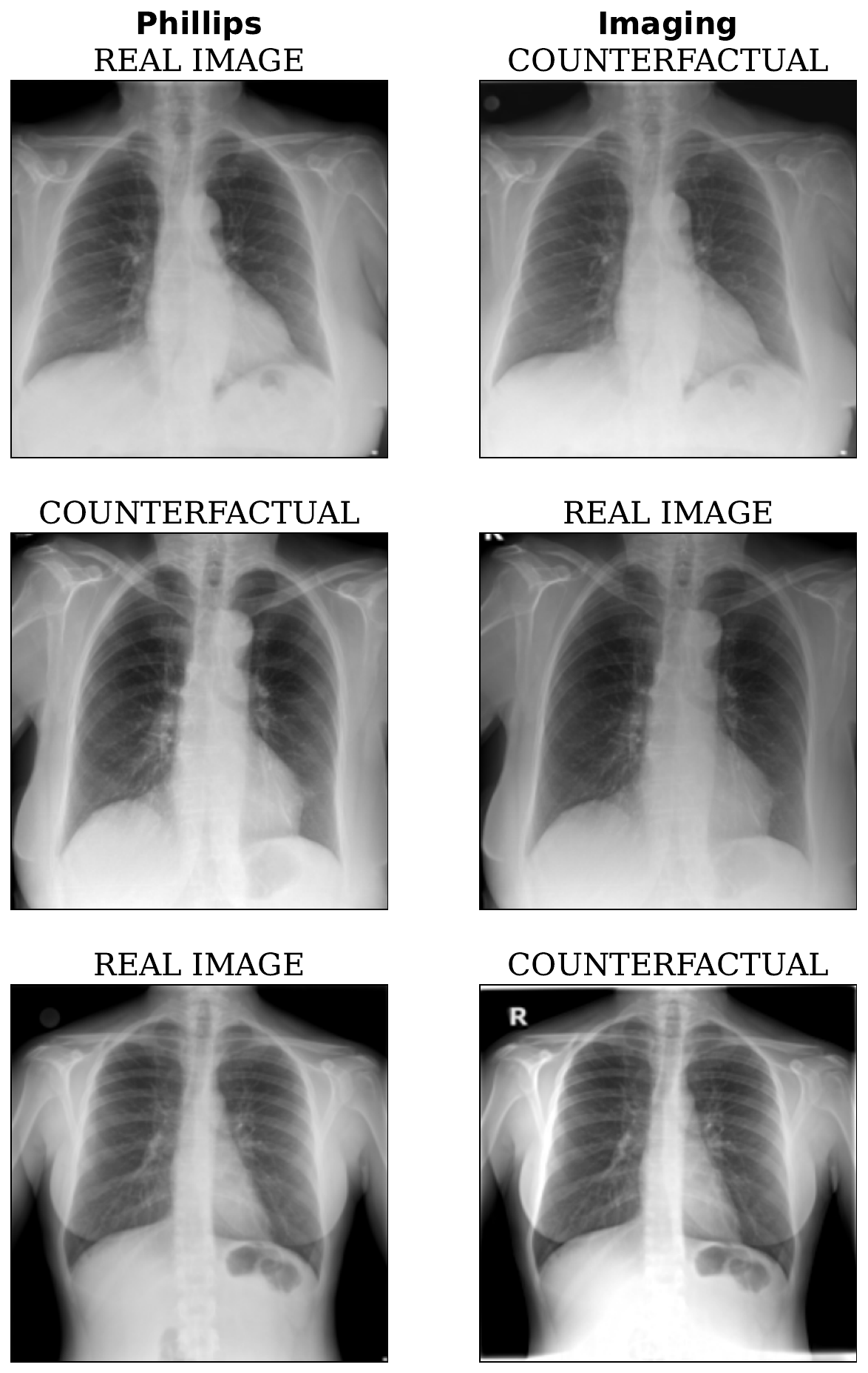}
    \caption{PadChest}
  \end{subfigure}
  \hfill
  \caption{\textbf{Examples of counterfactual images generated with our model}. Note that on PadChest, text is only imprinted on a subset of Imaging scans (not on Phillips): our model respects this by removing text when generating counterfactuals from Imaging to Phillips and vice-versa. Generated images have a resolution of 224x224 pixels for PadChest and 224x192 for EMBED.}
  \label{fig:cf_viz}
\end{figure}

To train the Deep Structural Causal Model, we need to specify the causal graph outlining the data-generating process. Most applications require the causal graph to closely describe true physiological and imaging processes affecting image appearance. However, in this study we only intervene on one variable, the `scanner' indicator. As such, we can simplify the causal graph to only contain this one variable, as factors of variations unaccounted for in the causal graphs will be captured by the exogenous noise. Importantly, with this minimalist graph, we do not need to condition the generation model on any downstream task labels, which is essential to preserve the unsupervised nature of the pretraining step. In our examples, we include scanner as the only parent in the causal graph for mammography generation and include both biological sex and scanner for chest radiography counterfactual inference (sex is optional for domain counterfactual generation). We provide a visual representation of these causal graphs in ~\cref{fig:causalgraphs}. For EMBED, we use weighted batch sampling during training to counter the imbalance in the scanner distribution. We provide qualitative examples of generated domain counterfactuals in \cref{fig:cf_viz}. In terms of counterfactual effectiveness~\citep{monteiro_measuring_2022}, generated scanner counterfactuals can deceive a scanner classifier trained on real data 94\% of the time for the PadChest model, and 77\% of the time for the EMBED model (when generating counterfactuals uniformly across domains). For sex counterfactuals on PadChest, we observed an effectiveness of 77\% .

\begin{figure}
\begin{subfigure}{\linewidth}
    \centering
    \includegraphics[width=\linewidth]{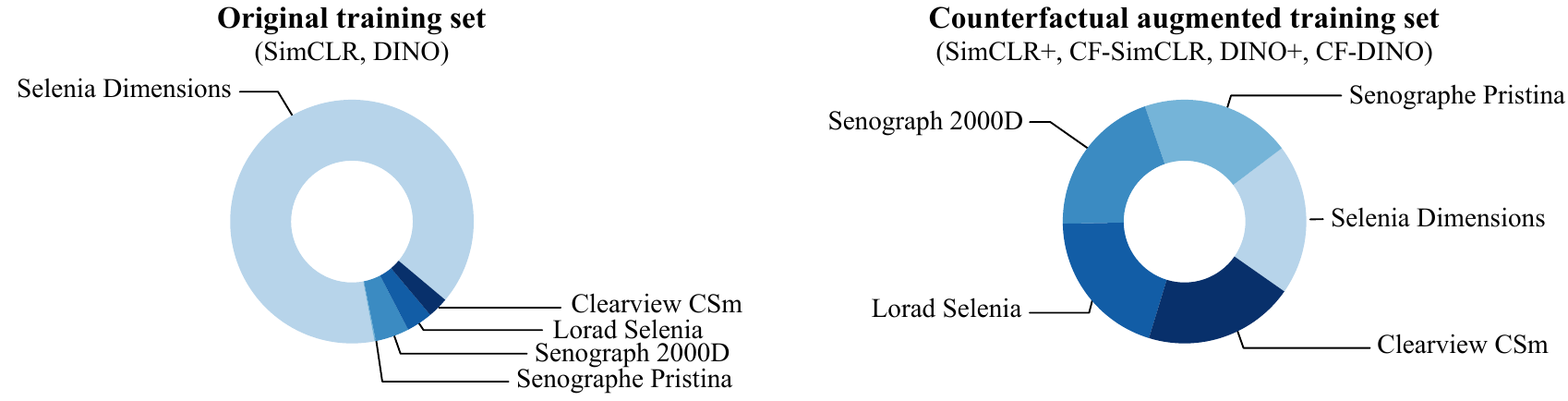}
    \caption{EMBED}
\end{subfigure}
\begin{subfigure}{\linewidth}
    \centering
    \includegraphics[width=\linewidth]{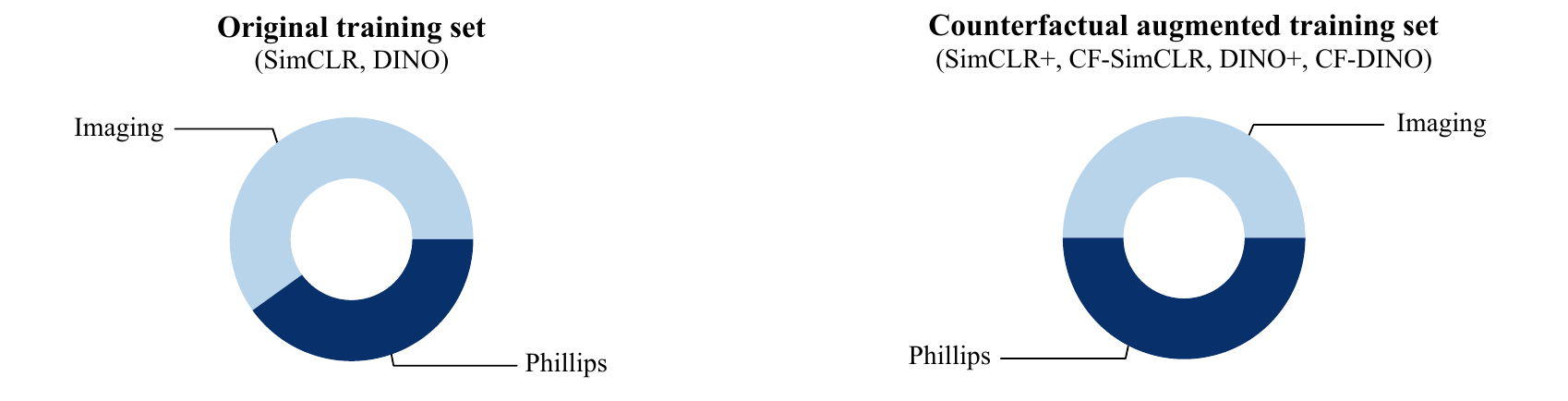}
    \caption{PadChest}
\end{subfigure}
\caption{\textbf{Distribution of scanners in the original (real-only) training set and the counterfactual-augmented training set for EMBED (top) and PadChest (bottom).}}
\label{fig:scanner_distribution}
\end{figure}

In CF-SimCLR, we construct positive pairs by combining a real image with a domain counterfactual, where the domain is randomly selected out of all possible domains available in the training set (see~\cref{sec:cfsimclr}). Specifically, we first generate all possible domain counterfactuals using all scanners at hand in the training set. That is, for EMBED, for each image, we generate four possible counterfactuals (all available scanners except the real one); for PadChest, we generate one counterfactual for each image (the other scanner). This means that the combined training set of real + generated counterfactuals is balanced across scanners. We use this extended training set to train both CF-SimCLR and SimCLR\textit{+} (resp. CF-DINO and DINO\textit{+}). The difference in scanner distribution between the original (real-only) training set and the counterfactual-augmented training set is depicted in \cref{fig:scanner_distribution}.

\subsection{Implementations details for self-supervised pretraining} 

We use ResNet-50~\citep{he_deep_2016} encoders (initialised with ImageNet weights) for all models pretrained with SimCLR. DINO-v2 use ViT-Base~\citep{dosovitskiy_image_2020} encoders, initialised with the weights from ImageNet DINO-v2. We used the original DINO-v2 code and hyperparameters~\citep{dosovitskiy_image_2020} for training and kept the encoder with the lowest validation loss. We kept all hyperparameters constant when comparing various contrastive pair generation strategies.

\section{Results}

In this section, we compare the quality and robustness of the learned representations for various pre-training paradigms. First, standard SimCLR. Secondly, SimCLR\texttt{+}, where we train a model using classic SimCLR on a training set enriched with domain counterfactuals. Finally, CF-SimCLR combines SimCLR with our proposed counterfactual contrastive pair generation framework. We then repeat the same analysis for models pre-trained with the DINO objective, comparing DINO, CF-DINO and DINO\texttt{+}. Note that in SimCLR\texttt{+} (resp. DINO\texttt{+}), counterfactuals and real images are not paired during the contrastive learning step; they are all considered as independent training samples. As such, SimCLR\texttt{+}/DINO\texttt{+} represent the common paradigm of simply enriching the training set with synthetic examples. In CF-SimCLR/CF-DINO, on the other hand, we systematically pair real images with their corresponding counterfactual for positive pair creation (\cref{fig:fig1}). 

We compare the effect of these three pretraining strategies on chest X-rays and mammograms. For chest X-rays, we evaluate the quality of the learned representations by assessing downstream performance on pneumonia detection. For mammography, we focus on the task of breast density prediction (important for risk modelling). Pre-training strategies are evaluated with linear probing (i.e. classifiers trained on top of frozen encoders) as well as full model finetuning (unfrozen encoders). Linear probes are best representative of the quality of learned representation during pre-training, as representations are unchanged during downstream training. However we also include the comparison with full model finetuning, as this is often the training paradigm of choice in practical scenarios. All models are finetuned with real data only, using a weighted cross-entropy loss. We evaluate the pre-trained encoders in two settings. First, we test the encoders on ID datasets, i.e. using the same data for pre-training, finetuning and testing. Secondly, encoders are evaluated on OOD datasets, i.e. where the model is finetuned/linear-probed and tested on data external to the pre-training data. Evaluation on external datasets is crucial to assess how counterfactual contrastive pretraining performs on unseen domains (outside of scanner distribution used for training the causal inference model). All encoders are pretrained on the full, unlabelled, PadChest and EMBED datasets. However, the main motivation for self-supervised pretraining is to increase robustness when only a limited amount of labelled data is available~\citep{azizi_robust_2023}. Hence, we evaluate the encoders for varying amounts of annotated data. Specifically, we finetune (or linear-probe) the encoder using a pre-defined amount of labelled samples, and then evaluate the resulting classifier on a fixed test set. We repeat this process several times, varying the amount of labelled samples to assess the effect of pre-training in function of number of labelled samples available for training the downstream classifier, e.g. for PadChest, the number of labelled training samples varies from 3,249 to 64,989. All our code is publicly available at \url{https://github.com/biomedia-mira/counterfactual-contrastive}.

For each task, we compare downstream performance across various pretraining strategies, for each scanner, for both SimCLR (\cref{fig:simclr_xray,fig:simclr_mammo}) and DINO (\cref{fig:dino_mammo,fig:dino_xray}) objectives.  Moreover, to help visualise performance differences across encoders, we report the performance differences between the proposed encoders and the baseline in \cref{fig:simclr_cxr_perf_difference,fig:simclr_mammo_perf_difference,fig:dino_cxr_perf_difference,fig:dino_mammo_perf_difference}.

\subsection{Does counterfactual contrastive learning improve performance and robustness under acquisition shift?}
\label{sec:cfsimclr:results}
First, we focus on comparing counterfactual contrastive strategies (CF-SimCLR, CF-DINO) versus standard contrastive learning (SimCLR, DINO) to assess whether our cross-domain contrastive pair improve downstream performance across various domains.

\begin{figure}
    \centering
    \includegraphics[width=\textwidth]{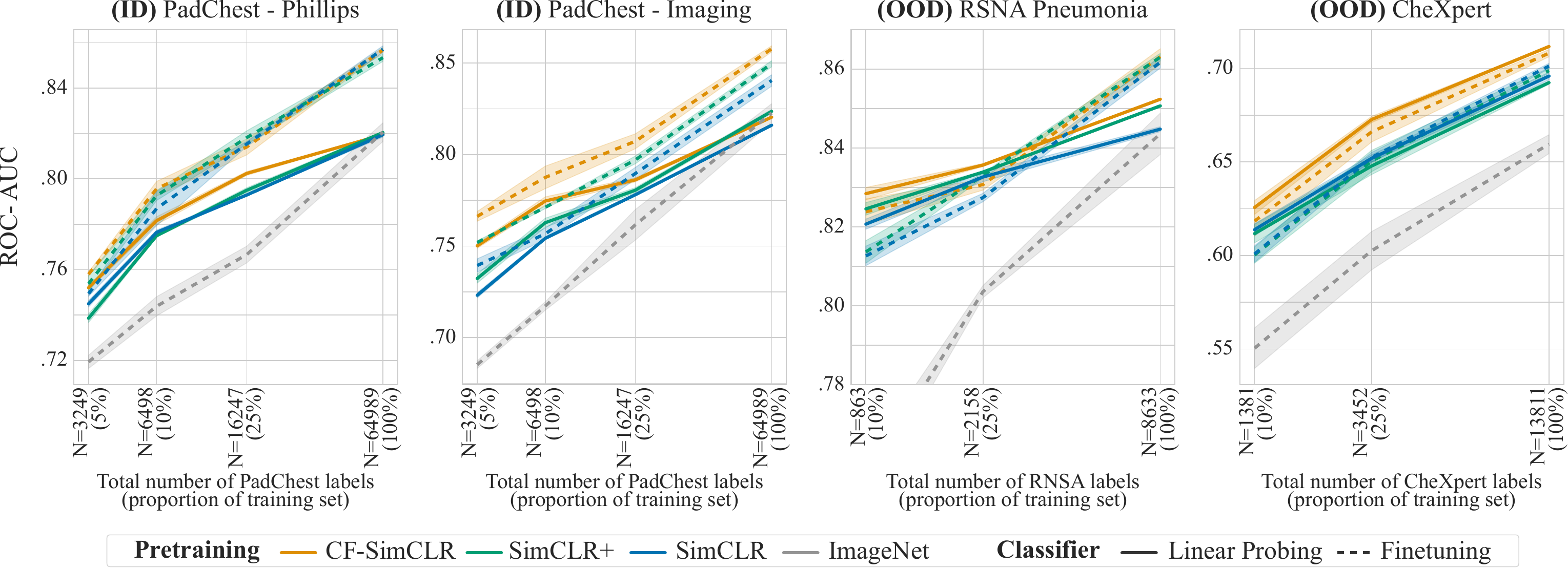}
    \caption{\textbf{Pneumonia detection results with linear probing  (frozen encoder, solid lines) and finetuning (unfrozen encoder, dashed lines) for models trained with the SimCLR objective}. Results are reported as average ROC-AUC over 3 seeds, shaded areas denote +/- one standard error. We also compare self-supervised encoders to a supervised baseline initialised with ImageNet weights.}
    \label{fig:simclr_xray}
\end{figure}

First, we focus on comparing counterfactual contrastive strategies (CF-SimCLR, CF-DINO) versus standard contrastive learning (SimCLR, DINO) to assess whether our cross-domain contrastive pair improve downstream performance across various domains (i.e. orange versus blue in \cref{fig:simclr_xray,fig:simclr_mammo} and orange bars in \cref{fig:simclr_cxr_perf_difference,fig:simclr_mammo_perf_difference}). We will compare CF-SimCLR with SimCLR\texttt{+} in the next section.

Results on pneumonia detection, in \cref{fig:simclr_xray}, show that CF-SimCLR outperforms the SimCLR baseline (orange versus blue), across datasets, irrespective of the amount of labels, with improvements particularly striking for linear probing. Gains can be best observed on the performance difference plots in \cref{fig:simclr_cxr_perf_difference}, where we observe that in all settings the performance difference compared to the SimCLR baseline is positive, i.e. CF-SimCLR performs better than SimCLR. For example, for linear probing, on PadChest, when using 3,249 training samples for training the Pneumonia classifier, the performance improves by 2.5\% on the Imaging scanner from PadChest and 0.6\% ROC-AUC on the Phillips scanner. On CheXpert, performance gains vary between 1\% and 2\% compared to the baseline with linear probing.   Mammography results in \cref{fig:simclr_mammo_perf_difference} show that CF-SimCLR consistently outperforms the SimCLR baseline across most ID scanners for both linear probing and finetuning, particularly when the amount of labelled data is limited (<20k). On the external VinDR dataset, CF-SimCLR beats both baselines on both scanners in the low data regime. CF-SimCLR pretraining mainly benefits scanners under-represented in the training set (all except Selenia Dimensions for EMBED and PlanMed Nuance for VinDr), regardless of the amount of labelled data. For example, in linear probing, for EMBED, with 2,230 labelled samples, performance increases by 3\% on Clearview CSM, 0.7\% on Lorad Selenia, 1.3\% on Pristina when comparing CF-SimCLR with SimCLR. Importantly, CF-SimCLR benefits the OOD scanners as well, in particular VinDR, where improvements range between 4\%-6\% ROCAUC compared to SimCLR when using 560-1,121 labelled samples. For higher levels of labels, the differences between encoders slowly vanish, particularly in finetuning settings. This is expected as starting representations matter less if many labels are available for training the classifier, during which learned representations may be substantially updated (in full model finetuning). For both modalities, the improvement on external datasets is particularly worth highlighting, given that the encoder was not exposed to these external domains during pretraining (nor during counterfactual generation).

Crucially, performance improvements and increased robustness to acquisition shift equally hold for encoders trained with the DINO-v2 objective, demonstrating the versatility of the proposed method. In \cref{fig:dino_mammo,fig:dino_mammo_perf_difference}, we can see that CF-DINO outperforms DINO for all EMBED scanners, across all levels of labels, for both linear probing and full model finetuning. Interestingly, on this dataset, we note that, for most scanners, the performance improvements between CF-DINO and DINO are even larger than between CF-SimCLR and SimCLR. Again, the gains mostly affect scanners under-represented during training with CF-DINO closing the performance gap between the majority scanner (Selenia Dimensions) and the other scanners (\cref{fig:dino_mammo}). For linear probes, in~\cref{fig:dino_mammo_perf_difference}, we note that the improvement for low levels of labels (<20k) are substantial for some under-represented scanners on EMBED: on Senograph 2000D (resp. Clearview CSm) we see a 4\% (resp. 2.8\%) improvement with 2,223 labelled samples and 2\% (resp. 1\%) with 11,154 labels. On Lorad Selenia, improvements vary between 1\% and 1.5\% when training with up to 55k samples, whereas on Senographe Pristina, we observe 1\% improvements with 2,223 labelled samples and 2\% improvements for all other levels of labels. For VinDR, as for CF-SimCLR, performance improvements are mostly noticeable for the minority scanner PlanMed nuance, where performance gains are 5\% for 560 labels, 6\% for 1,121 labels and 2\% for 2,803 labels, with gains vanishing for 11,212 labels (where all encoders perform equally). For finetuning, performance gains are mostly observed for low levels of labels and under-represented scanners (bottom two rows in~\cref{fig:dino_mammo_perf_difference}). For chest X-rays, in \cref{fig:dino_xray,fig:dino_cxr_perf_difference}, we can see that training with CF-DINO closes the performance gap between both PadChest scanners for both linear probing evaluation and finetuning. Indeed, without CF-DINO, there is a substantial performance drop from Phillips images to Imaging images, whereas this gap is much smaller with performance improving substantially with CF-DINO across all levels of labels for images from Imaging (\cref{fig:dino_xray}). Similarly, in \cref{fig:dino_cxr_perf_difference}, we observed that CF-DINO outperforms DINO on the external RSNA Pneumonia dataset by a substantial margin 2\% with 863 labelled samples, 3.5\% with 2,158 labels and 1.1\% with the full training set with linear probe, and improvements varying from 1 to 2\% in finetuning as well. For CheXpert, CF-DINO outperforms DINO in model finetuning (and linear probing with 10\% of labels). However, it slightly underperforms on the 25\% and 100\% for linear probing. Note that results on the CheXpert dataset are to be interpreted with some caution as labels are NLP sourced and, in general, of rather low quality~\citep{irvin_chexpert_2019}, which may explain the consequent drop of performance for all models between CheXpert and the expert-labelled RSNA Pneumonia dataset.

\subsection{What is the benefit of counterfactual contrastive learning over simply extending the training set with the generated counterfactual data?}
\label{sec:results:simclrplus}
We have shown that the models trained with counterfactual contrastive learning outperform models trained with standard contrastive pairs. However, it is important to note that models trained with CF-SimCLR (resp. CF-DINO), are exposed to additional (synthetic) data during training. Hence, in this section, to isolate the effect of the `smart pair creation' proposed in this work, we compare it to another baseline, SimCLR\texttt{+} (resp. DINO\texttt{+}), where we use the same extended training set but where generated samples are considered as independent samples and are not paired with real images during training, details can be found in \cref{sec:cfsimclr}. 

\begin{figure}
     \centering
    \includegraphics[width=\textwidth]{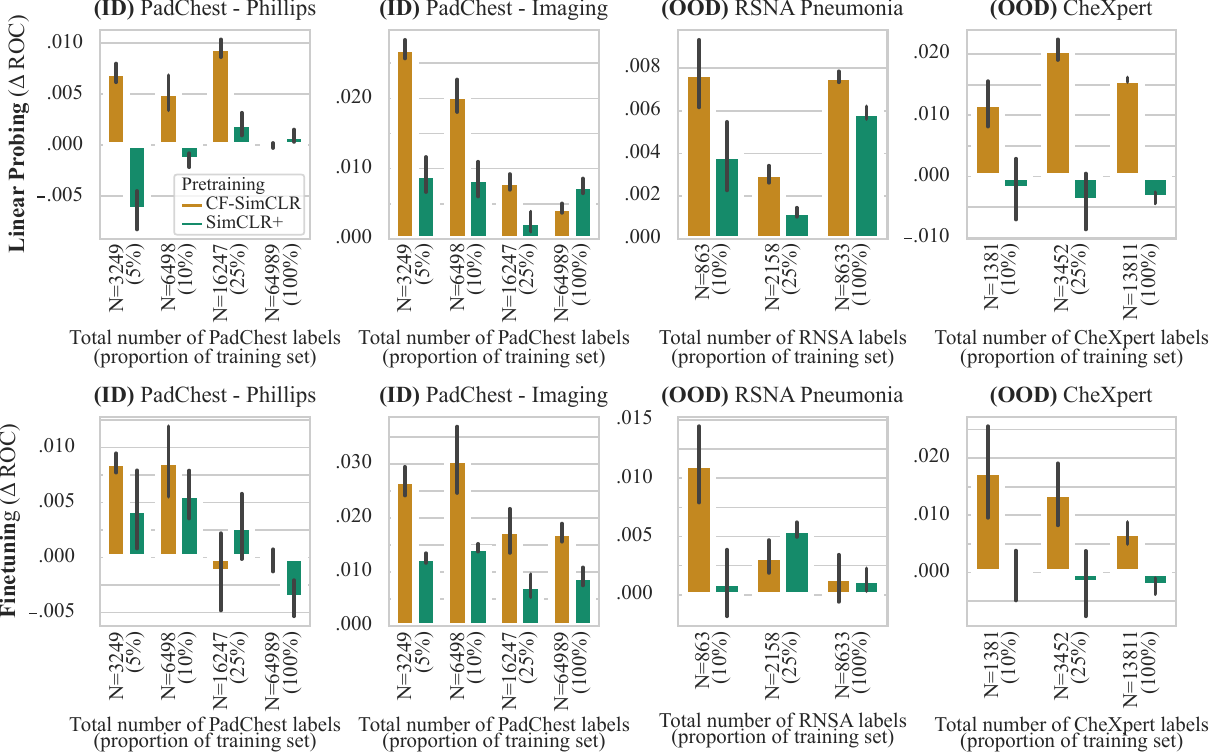}
     \caption{\textbf{ROC-AUC difference to SimCLR baseline for CF-SimCLR and SimCLR\texttt{+} for pneumonia detection}. The top row depicts results with linear probing, bottom row shows results with model finetuning. Results are reported as average ROC-AUC difference compared to the baseline (SimCLR) over 3 seeds, error bars denote +/- one standard error. CF-SimCLR consistently outperforms encoders trained with standard SimCLR and SimCLR\texttt{+} (where counterfactuals are added to the training set) for linear probing, and performs best overall for full model finetuning.}
     \label{fig:simclr_cxr_perf_difference}
 \end{figure}

\begin{figure}
    \centering
    \includegraphics[width=\textwidth]{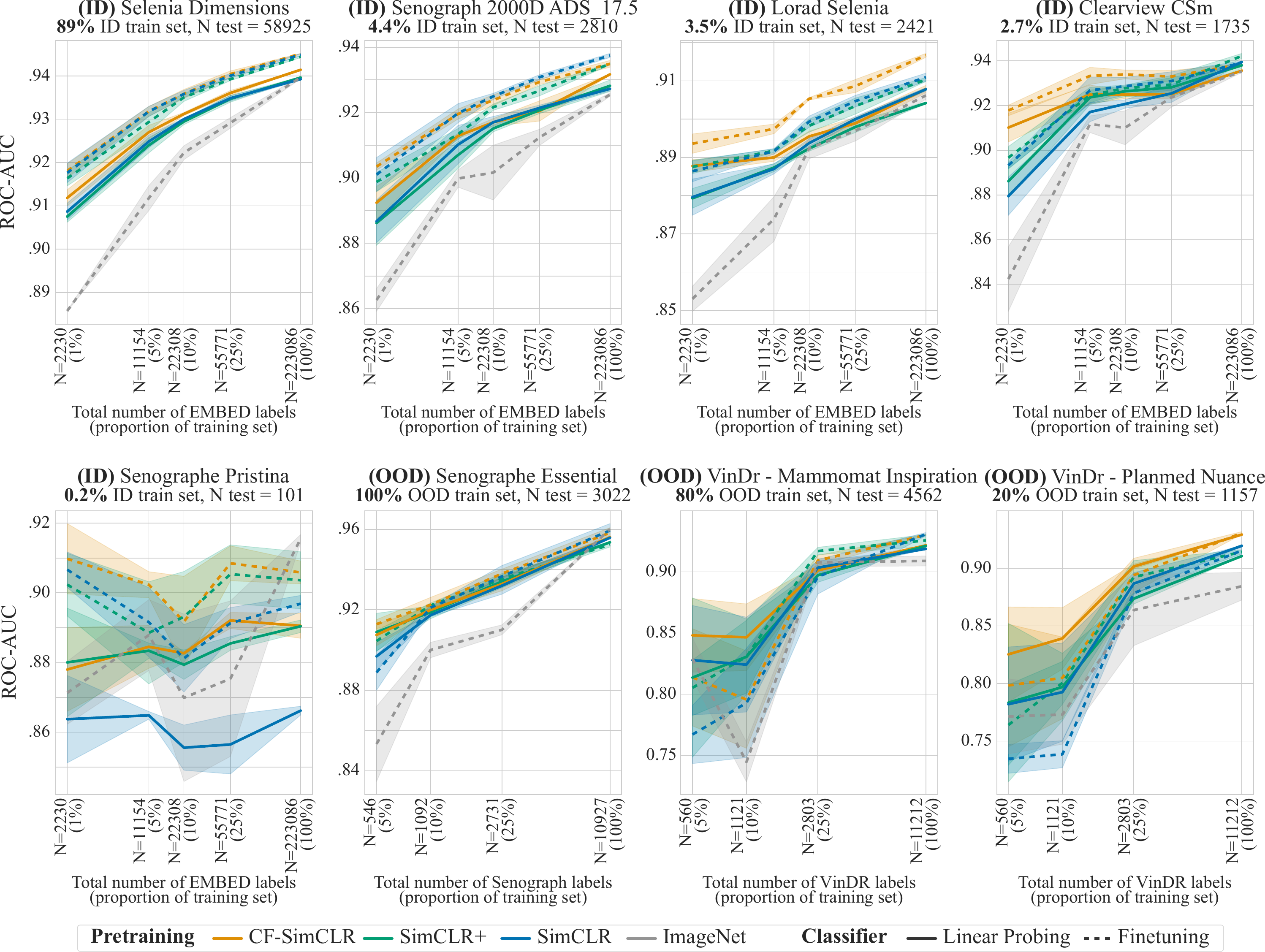}
    \caption{\textbf{Breast density results with linear probing (frozen encoder, solid lines) and finetuning (unfrozen encoder, dashed lines) for models trained with SimCLR}. Results are reported as average one-versus-rest macro ROC-AUC over 3 seeds, shaded areas denote +/- one standard error. CF-SimCLR performs best overall across ID and OOD data, and improvements are largest in the low data regime and on under-represented scanners.}
    \label{fig:simclr_mammo}
\end{figure}

\begin{figure}
     \centering
    \includegraphics[width=\textwidth]{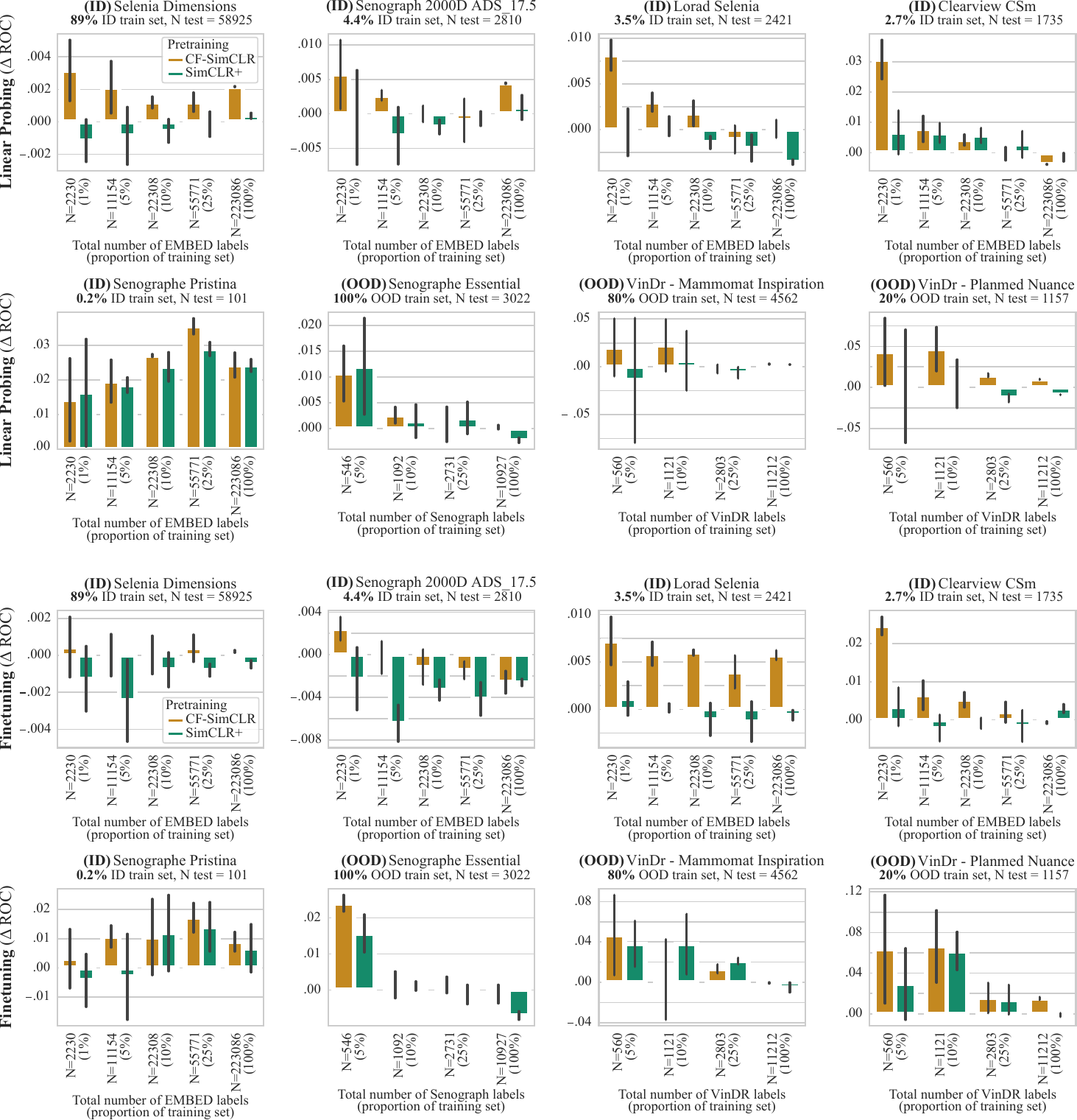}
     \caption{\textbf{ROC-AUC difference between SimCLR and CF-SimCLR (resp. SimCLR\texttt{+}) for breast density assessment}. The top two rows denote results with linear probing, and the bottom two rows show results with model finetuning. Results are reported as average macro ROC-AUC difference compared to the baseline (SimCLR) over 3 seeds, error bars denote +/- one standard error. CF-SimCLR overall performs best across ID and OOD data, improvements are largest in the low data regime and on under-represented scanners.}
     \label{fig:simclr_mammo_perf_difference}
 \end{figure}

\begin{figure}
    \centering
    \includegraphics[width=\textwidth]{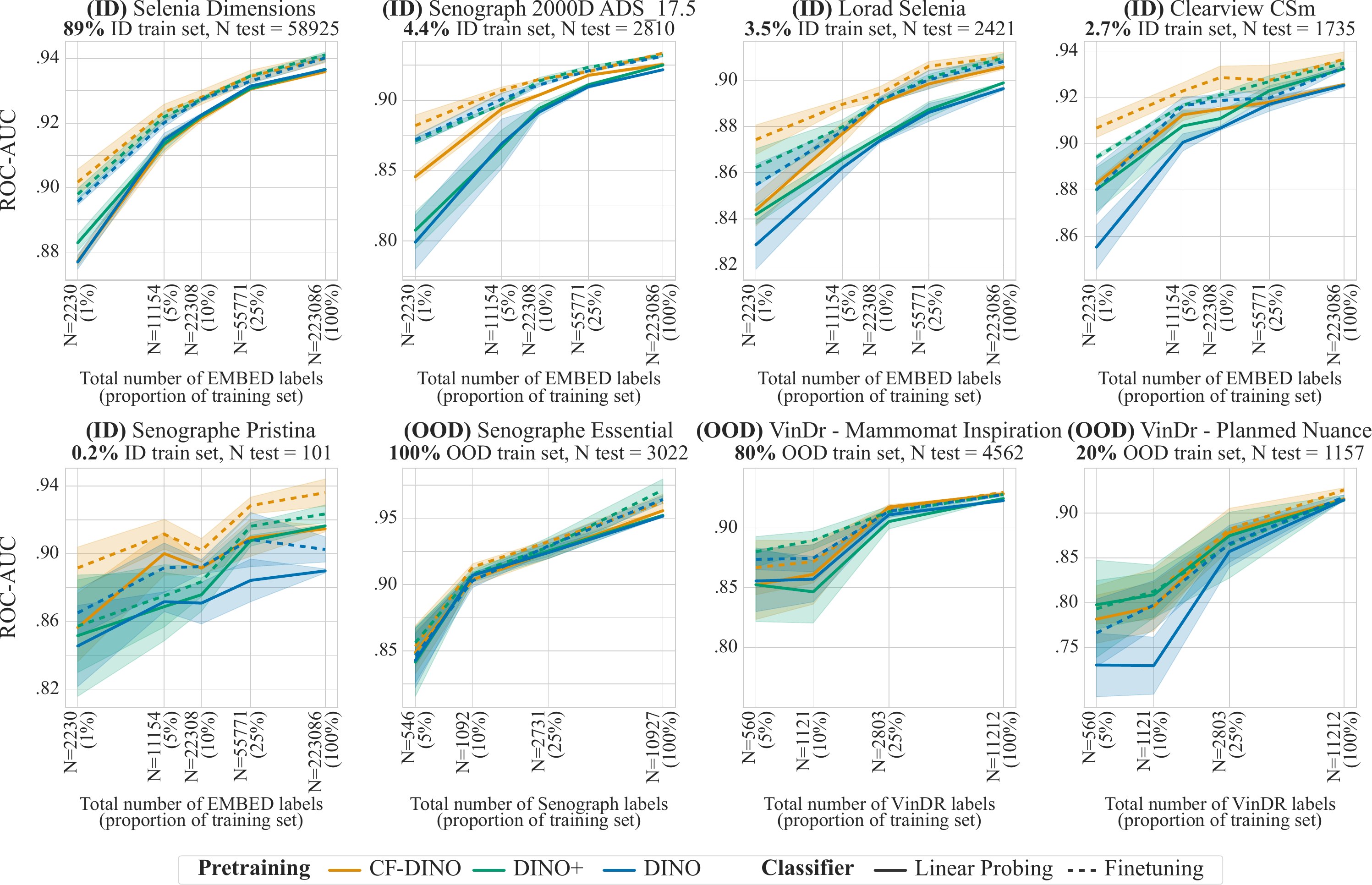}
    \caption{\textbf{Breast density classification results for models pretrained with DINO-v2, for both linear probing and finetuning}. Results are reported as average one-versus-rest macro ROC-AUC over 3 seeds, shaded areas denote +/- one standard error. CF-DINO performs best overall, across ID and OOD data, improvements are largest in the low data regime.}
    \label{fig:dino_mammo}
\end{figure}

\begin{figure}
     \centering
    \includegraphics[width=\textwidth]{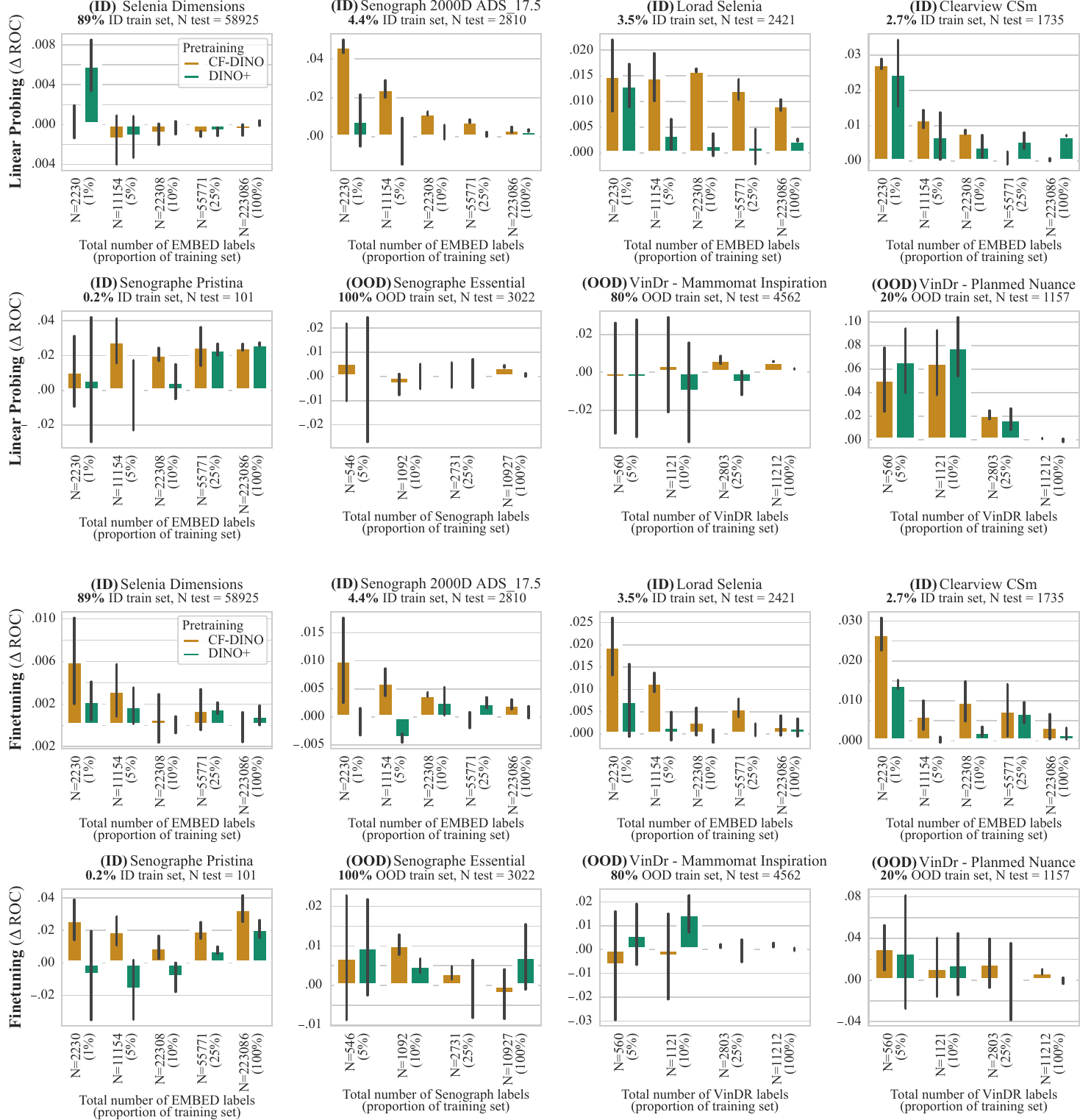}
     \caption{\textbf{ROC-AUC difference between DINO and CF-DINO (resp. DINO\texttt{+})}. Top two rows denote results with linear probing, bottom two rows results with model finetuning. Results are reported as average macro ROC-AUC difference compared to the baseline (DINO) over 3 seeds, error bars denote +/- one standard error. CF-DINO overall performs best across ID and OOD data, improvements are largest in the low data regime and on under-represented scanners.}
     \label{fig:dino_mammo_perf_difference}
 \end{figure}

\begin{figure}
    \centering
    \includegraphics[width=\textwidth]{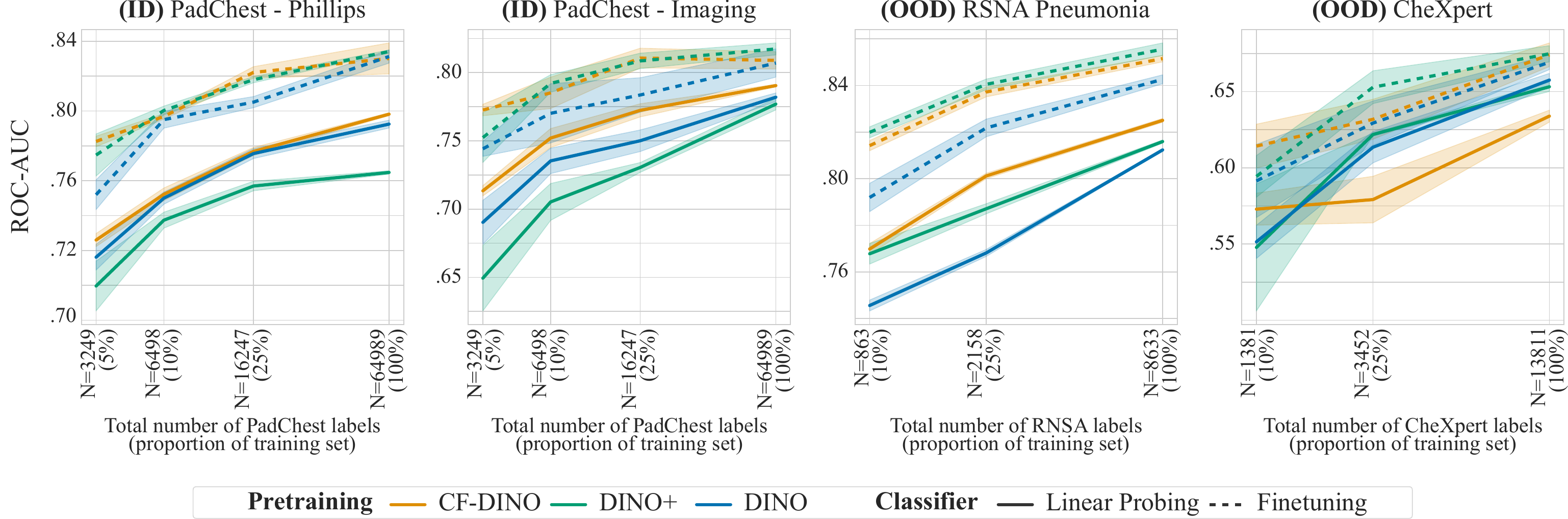}
    \caption{\textbf{Pneumonia detection results for models trained with DINO-v2, for both linear probing (frozen encoder) and finetuning}. Results are reported as average ROC-AUC over 3 seeds, shaded areas denote +/- one standard error. CF-DINO consistently outperforms standard DINO.}
    \label{fig:dino_xray}
\end{figure}

\begin{figure}
     \centering
    \includegraphics[width=\textwidth]{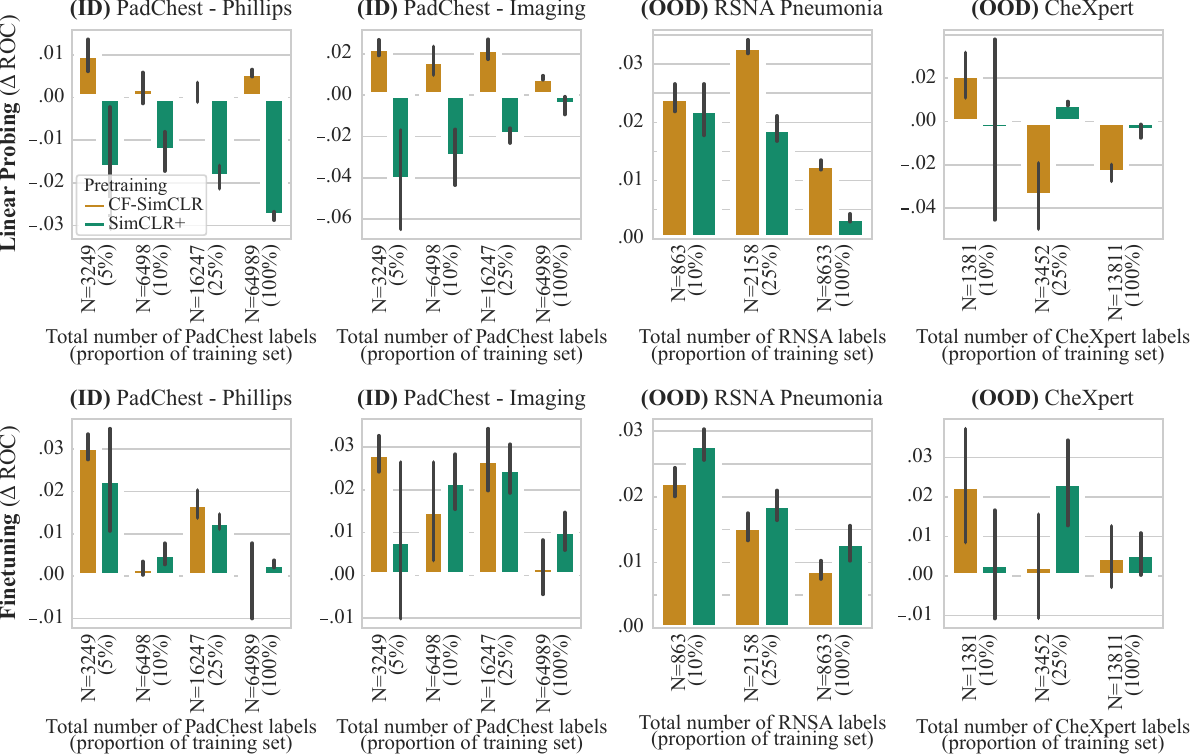}
     \caption{\textbf{ROC-AUC difference to DINO baseline for CF-DINO and DINO\texttt{+} for pneumonia detection}. The top row depicts results with linear probing, bottom row show results with model finetuning. Results are reported as average ROC-AUC difference compared to the baseline (DINO) over 3 seeds, error bars denote +/- one standard error.}
     \label{fig:dino_cxr_perf_difference}
 \end{figure}

Differences between CF-SimCLR and SimCLR\texttt{+} are best observed in \cref{fig:simclr_cxr_perf_difference,fig:simclr_mammo_perf_difference} (resp. \cref{fig:dino_cxr_perf_difference,fig:dino_mammo_perf_difference} for CF-DINO versus DINO\texttt{+}). Overall, CF-SimCLR outperformed SimCLR\texttt{+} consistently across all experimental settings (\cref{fig:simclr_mammo_perf_difference,fig:dino_mammo_perf_difference}), with similar results for CF-DINO. On mammography, CF-SimCLR/ outperformed SimCLR/DINO\texttt{+} across all EMBED datasets. CF-SimCLR consistently also outperforms SimCLR\texttt{+} on VinDr, while CF-DINO/DINO\texttt{+} perform similarly on VinDR. On chest X-rays, in \cref{fig:simclr_cxr_perf_difference}, CF-SimCLR consistently outperforms SimCLR\texttt{+} across all settings. In \cref{fig:dino_cxr_perf_difference}, CF-DINO outperformed DINO\texttt{+} by a large margin on PadChest and RSNA Pneumonia for linear probing, and both performed similarly for model finetuning. We especially noticed that the performance gains of SimCLR\texttt{+} (resp. DINO\texttt{+}) over SimCLR were not very stable, improving for some domains while performing on par with standard SimCLR (resp. DINO) for others. In general, the counterfactual contrastive approaches offered much more consistent performance improvements. The starkest differences were observed on scanners under-represented during training (see \cref{fig:simclr_mammo_perf_difference}). 

\begin{figure}
    \centering
    \includegraphics[width=\textwidth]{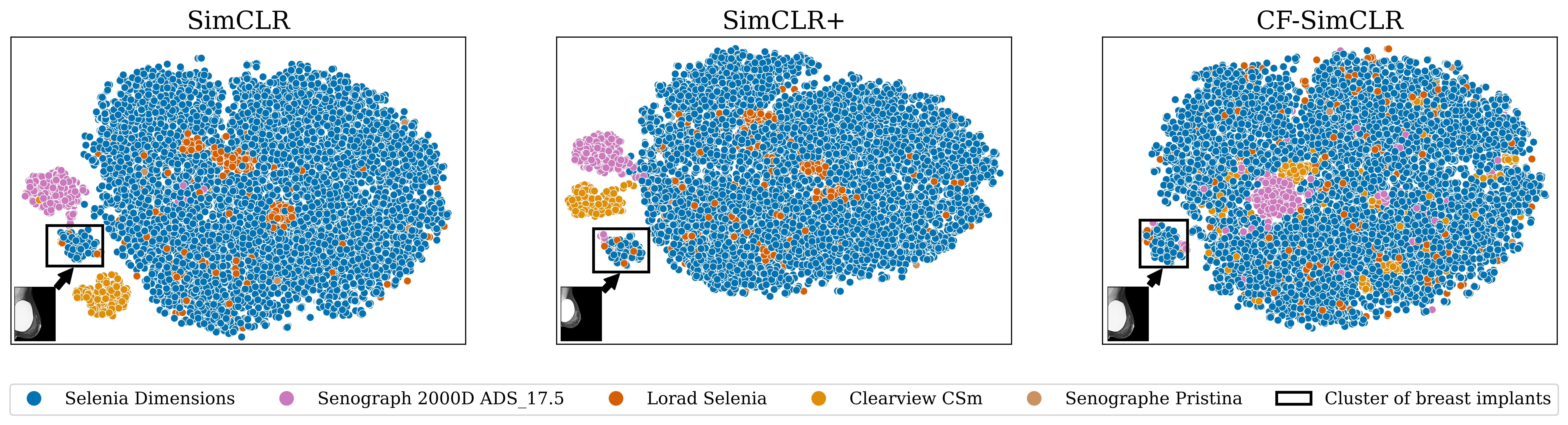}
    \caption{\textbf{t-SNE projections of embeddings from 16,000 randomly sampled test images from mammography encoders trained with SimCLR, SimCLR+ and CF-SimCLR}. Encoders trained with SimCLR and SimCLR\texttt{+} exhibit domain clustering. CF-SimCLR embeddings are substantially less domain-separated and the only disjoint cluster exclusively contains breasts with implants, semantically different. Thumbnails show a randomly sampled image from each `implant' cluster. Adapted from \citep{conferencepaper}.}
    \label{fig:mammo-embeddings}
\end{figure}

These experimental results align with the theoretical benefits of counterfactual contrastive learning: by explicitly creating realistic cross-domain positive pairs during training, we directly encourage the network to create domain-agnostic image representations. The more domain-agnostic image representations are, the bigger the expected improvement in terms of robustness to acquisition shift. Such an increase in robustness leads to higher performance on devices under-represented during training, even more so in limited data settings. The increase in domain alignment is directly visible in the t-SNE plots of feature embeddings in \cref{fig:mammo-embeddings}, where embeddings of mammography models trained with SimCLR and SimCLR\texttt{+} exhibit clear domain clustering, whereas the model trained with CF-SimCLR generate embeddings where domains are less separated. To further assess the domain separability of the features from those encoders, we train a scanner classifier on the features from each classifier. To reduce scanner-imbalance during training of the scanner classifier, we randomly select 1,000 images from the test set for each scanner (except for Senographe Pristina where we only had 101 test images), and extract corresponding features with each encoder. We then run 5-fold cross-validation, classifying scanners from the features of each pretrained model, using PCA (with 16 components) followed by logistic regression. Our results show that SimCLR and SimCLR+ achieve the same domain classification performance of 85\% and 87\% balanced accuracy, on the contrary CF-SimCLR only achieves 68\% balanced accuracy, confirming that CF-SimCLR significantly reduces scanner separability in feature space.

\subsection{Computational considerations} 

To implement counterfactual contrastive learning, we need to train an additional causal image generation model, this raises the question of computational overhead. Fortunately, the HVAE used here to generate image counterfactuals is relatively lightweight (as opposed to alternative generative approaches such as diffusion models). This HVAE not only trains relatively fast (e.g. only 20 epochs needed for EMBED, about 250k steps), it is also relatively frugal in terms of VRAM requirements (20GB of GPU VRAM were sufficient), which has the advantage of low hardware requirements. In terms of generation speed, we were able to generate over 1 million 224x224 mammography images in under 7 hours (on an NVIDIA RTX-3090 GPU). These computational requirements need to be put in perspective compared to the requirements of the contrastive pretraining step. Contrastive pretraining is resource intensive both in terms of VRAM (large batch sizes) and in terms of training time. Each SimCLR model was trained for 450 epochs for EMBED (resp. 1,000 epochs for PadChest), and required 2x46GB VRAM. For DINO, memory requirements were even higher (6x46GB VRAM for a batch size of 300). The overhead of the counterfactual generation part is negligible, and its benefits clearly outweigh the added computational costs.

\subsection{Ablation study on the impact of counterfactual quality on downstream performance}
\label{sec:ablationstudy}

In this section, we aim to assess the effect of the counterfactual image generation model quality on the downstream performance of the proposed counterfactual contrastive objective. To this end, we here compare three counterfactual image generation models, trained on EMBED:
\begin{itemize}
    \item \emph{HVAE\texttt{-}}: an HVAE-based counterfactual image generation as per~\citet{ribeiro_high_2023} trained for one epoch only (13k steps).
    \item \emph{HVAE}: the HVAE used in the experiment so far, trained for 20 epochs, without counterfactual finetuning.
    \item \emph{HVAE\texttt{+}FT}: where we additionally add the counterfactual finetuning step, as proposed by~\citet{ribeiro_high_2023}, to the \emph{HVAE} model. This step allows for improved effectiveness by further finetuning the counterfactual generation models using guidance from external classifiers.
\end{itemize}

\begin{figure}
    \centering
    \includegraphics[width=0.95\linewidth]{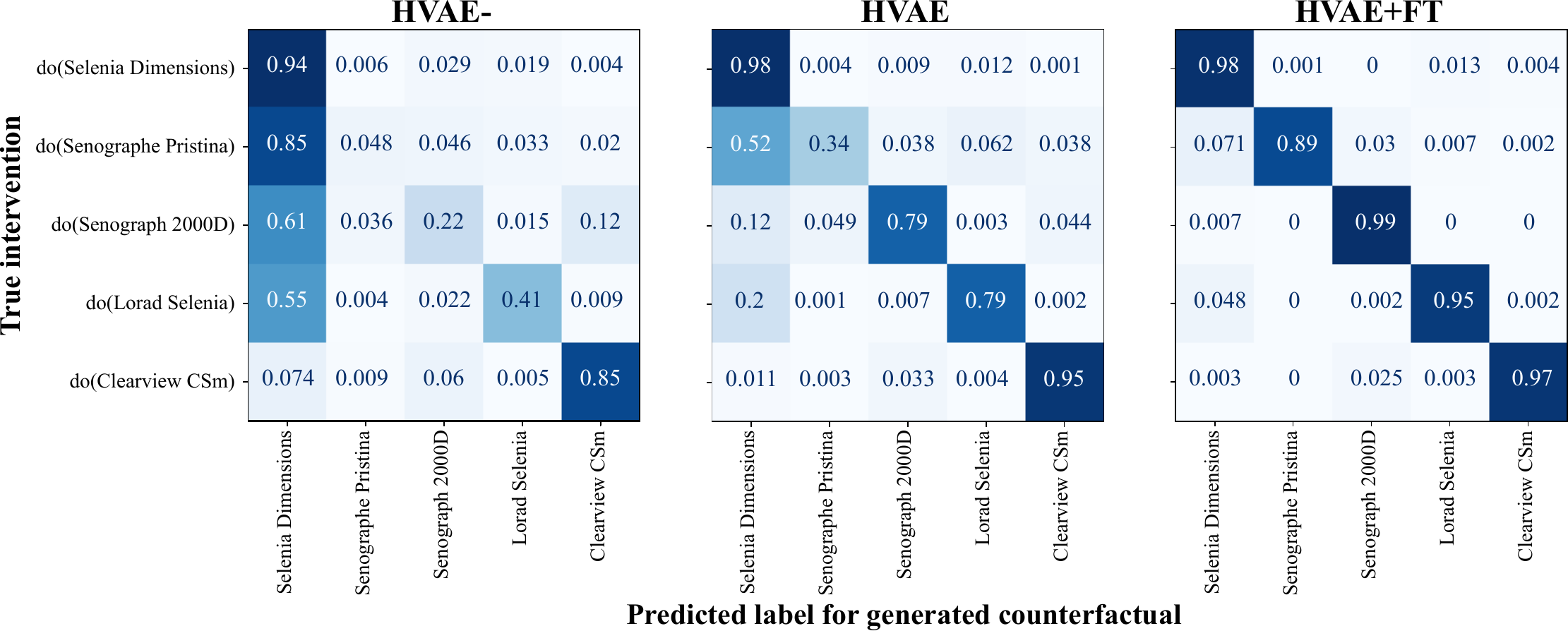}
    \caption{\textbf{Effectiveness comparison for the three counterfactual models considered in this ablation study}, by intervention. Computed on 8,304 validation set samples.}
    \label{fig:effectiveness}
\end{figure}

\begin{figure}
    \centering
    \includegraphics[width=0.8\linewidth]{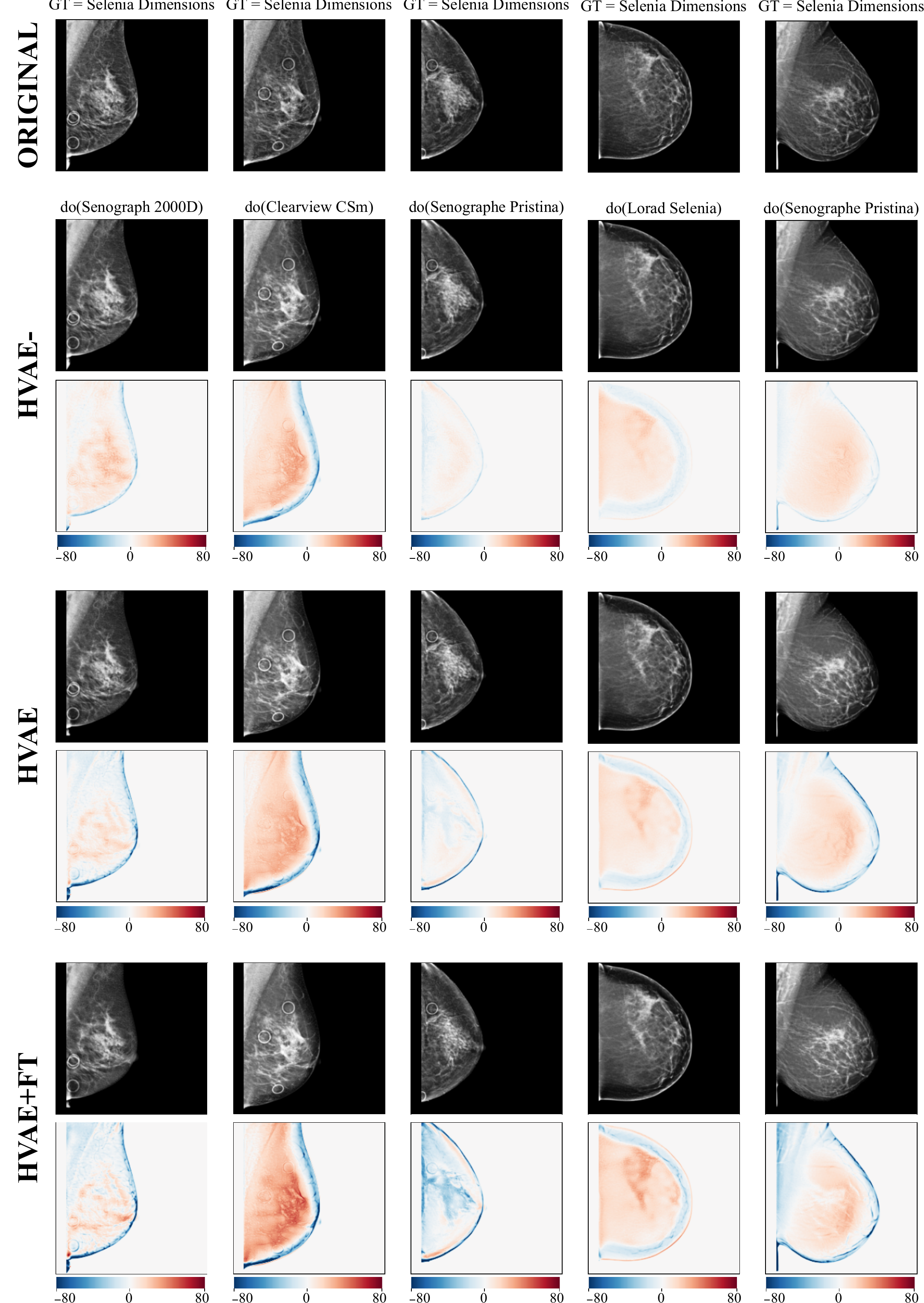}
    \caption{\textbf{Qualitative of comparison of the three counterfactual generation models, HVAE\texttt{-}, HVAE and HVAE\texttt{+}FT compared in the ablation study}. For each model we show generated counterfactuals as well as direct effect maps. Direct effects give a visual depiction of the increase in effectiveness across the three models from top to bottom. We also observe that all models preserve semantic identity very well, a key aspect in positive pair creation contrastive learning.}
    \label{fig:quali-comparison}
\end{figure}

First, we assess counterfactual quality for all three models using the metrics proposed in~\citet{monteiro_measuring_2022}. Results in~\cref{fig:effectiveness} show that effectiveness varies strongly across models, especially for scanners under-represented in the training set (e.g. Senographe Pristina). Moreover, results in~\cref{tab:cf_metrics} show that while effectiveness varies strongly across the compared models, reversibility and composition are strong for all models, indicating that all models preserve identity well. This can also be seen in the qualitative examples in \cref{fig:quali-comparison}. Strong semantic identity preservation is key for creating meaningful contrastive pairs, as the network will consider any changes across views in a positive pair as non-semantic information to disregard. Besides computational considerations, this further motivates the use of the HVAE counterfactual approach in this study, over, for example, diffusion-based counterfactual generation approaches as diffusion models have been shown to not always preserve identity well~\citep{preechakul_diffusion_2022,mokady_null-text_2023}.

\begin{table}
    \centering
    \begin{tabular}{lccc}
    \toprule
      CF model   & $\mathrm{Effectiveness}$ &  $\mathrm{Reversibility}^{(1)}$ & $\mathrm{Composition}^{(1)}$ \\
         & do(scanner) &  \\
        \midrule
        HVAE\texttt{-} & 49\% & 0.001 & 8e-12 \\
        HVAE & 77\% & 0.002 & 5e-12 \\
        HVAE\texttt{+}FT & 96\% & 0.004  & 5e-12\\
    \bottomrule
    \end{tabular}
    \caption{\textbf{Axiomatic soundness metrics~\citep{monteiro_measuring_2022} for the three models considered in this ablation study}. Computed on the 8,304 validation set samples.}
    \label{tab:cf_metrics}
\end{table}

\begin{figure}
    \centering
    \includegraphics[width=\linewidth]{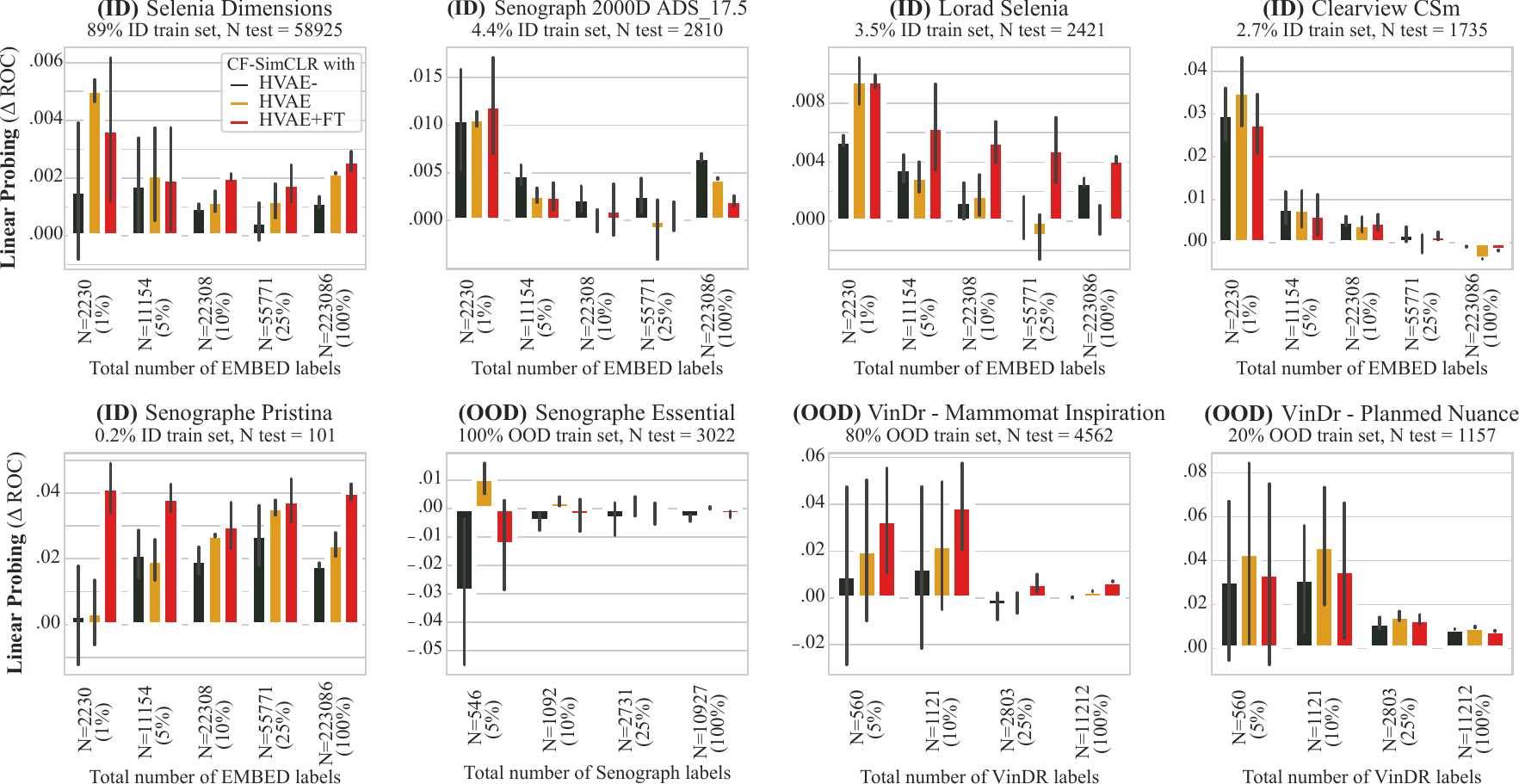}
    \caption{\textbf{Effect of counterfactual quality on downstream performance}. Results are reported as average macro ROC-AUC difference compared to the baseline (SimCLR) over 3 seeds for linear probing, error bars denote +/- one standard error. We compare running CF-SimCLR with (i) HVAE\texttt{-} a counterfactual generation model of lesser effectiveness, (ii) HVAE the generation model used in the rest of this study, (iii) HVAE\texttt{+}FT a counterfactual generation model with higher effectiveness.}
    \label{fig:ablation_counterfactual_effectiveness}
\end{figure}

In~\cref{fig:ablation_counterfactual_effectiveness}, we compare the downstream performance of the three CF-SimCLR encoders trained using generated counterfactuals from the three counterfactual generation models analysed above: HVAE\texttt{-}, HVAE and HVAE\texttt{+}FT. The graphs depict differences in ROC performance compared to the SimCLR baseline (i.e. no counterfactuals) across several levels on labels and scanners. These results first show that even when using a model with relatively low effectiveness, CF-SimCLR with HVAE\texttt{-} is either on par with or outperforms the baseline (positive performance differences), especially on under-represented scanners and for lower levels of labels. However, as effectiveness increases, the downstream performance for under-represented scanners tends to increase. We can see that the effectiveness of the HVAE model is substantially better compared to the HVAE\texttt{-} model for generating Lorad Selenia (+30\%) and Senographe Pristina (+70\%) counterfactuals. Further improving effectiveness with counterfactual finetuning (i.e. HVAE\texttt{+}FT) leads to striking improvements in the most under-represented scanner Senograph Pristina (where effectiveness increases by 60\% between HVAE and HVAE\texttt{+}FT). On this scanner, we see improvements of 4\%, 2\%, 0.5\% and 1.8\% ROC-AUC between CF-SimCLR with HVAE and CF-SimCLR with HVAE\texttt{+}FT across the various levels of labels, while also observing noticeable improvements on ID Lorad Selenia and the OOD VinDr Mammomat scanner. However, we also do notice somewhat diminishing returns of effectiveness improvements, where performance on ClearviewCSM, for example, is not substantially changing between CF-SimCLR with HVAE\texttt{-}, HVAEand HVAE\texttt{+}FT despite effectiveness improving from 85\% to 97\%, with similar results observed on Senograph 2000D where increase in effectiveness did not help to improve downstream performance. Results indicate that a scanner effectiveness above 80\% is satisfactory for CF-SimCLR.

\subsection{Robustness beyond acquisition shifts}
\label{sec:fair_cf}

\begin{figure}
    \centering
    \includegraphics[width=0.85\textwidth]{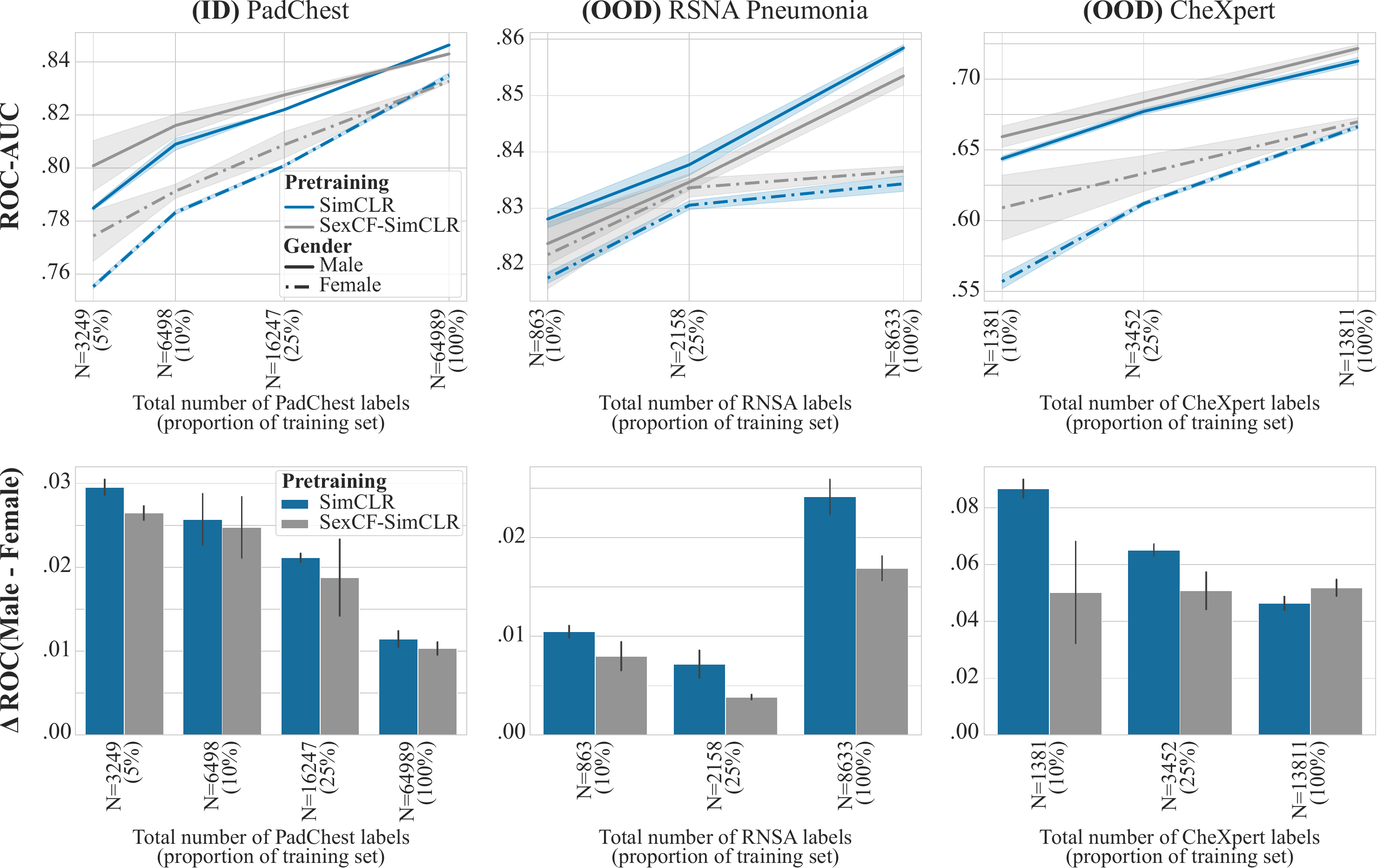}
    \caption{\textbf{Improving sub-group performance with counterfactual contrastive learning}. Pneumonia detection results with linear probing for encoders trained with SimCLR and SexCF-SimCLR. In SexCF-SimCLR, we generate sex counterfactuals instead of domain counterfactuals for positive pair generation to improve robustness to subgroup shift and, ultimately, performance on under-represented subgroups. Top row: performance for the male (solid line) and female (dashed line) subgroups, reported as average ROC-AUC over 3 seeds, shaded areas denote +/- standard error. Bottom row: performance disparities across the two subgroups, reported as average ROC-AUC difference between the male and female subgroups, over 3 seeds. Sex CF-SimCLR reduces sub-group disparities for all datasets, substantially increasing performance on the female sub-group when limited amounts of labels are available, both on ID and OOD datasets.}
    \label{fig:fair_linear}
\end{figure}

The main focus of this work is to improve robustness to acquisition shift. However, by simply changing the types of generated counterfactuals, the counterfactual contrastive pair generation framework can be easily extended to other distribution shifts. For example, we may use the same framework to increase robustness to population shift, improving the performance of under-performing subgroups. We illustrate this on chest radiography in \cref{fig:fair_linear}, where we generate sex counterfactuals instead of domain counterfactuals in the positive pair creation step, aiming to reduce performance disparities between the `male' and `female' subgroups. We can see that the baseline approach (SimCLR) performs sub-optimally on female patients for pneumonia detection across all datasets (top row in \cref{fig:fair_linear}). Using cross-subgroup counterfactuals positive pairs by intervening on biological sex, we observe performance improvements across subgroups in all datasets and for most levels of training labels. For example, on PadChest we observe an improvement of 2\% ROC-AUC with 3249 labelled samples for females compared to the SimCLR baseline (resp. 1\% with 16247 labels). On CheXpert, we also get +5\% ROC-AUC with 1381 labelled samples, and +3\% with 3452 labels. For PadChest and CheXpert, we even observe small performance improvements on the male subgroup. On RSNA however, while Sex-CFSimCLR successfully reduces the performance gap across both groups (bottom row), this here comes at the cost of a slight performance reduction in the male subgroup (top row), a phenomenon commonly known as the `levelling down' effect in the fairness literature~\citep{zietlow_leveling_2022}. Overall, analysing performance differences across subgroups (bottom row in \cref{fig:fair_linear}), we notice that Sex-SimCLR yields a consistent reduction in performance disparities across all datasets. Like in previous experiments, we notice that performance improvements are most notable in settings where it is most beneficial to have aligned representations across subgroups: in limited label settings and for the underperforming subgroups.

\section{Discussion and Conclusions}
In this work, we present \emph{counterfactual contrastive learning}, a novel contrastive pair generation framework enhancing the robustness of contrastively-learned image representations to domain shifts. Evaluating across five datasets, two modalities and two clinically-relevant classification tasks, we show that the proposed counterfactual contrastive pretraining approach yields higher downstream performance than standard contrastive pretraining, improvements which are particularly noticeable for domains less represented in the training set. Moreover, the proposed counterfactual pair generation method is agnostic to the choice of the contrastive objective, as demonstrated by our experiments showing that counterfactual positive pair generation improves results for models using SimCLR as well as DINO-v2 objectives. Importantly, we show that CF-SimCLR (resp. CF-DINO) also improves performance on external domains, which are not included in the pretraining set compared to SimCLR (resp. DINO). Our experiment on subgroup counterfactual contrastive learning demonstrates its broader applicability beyond acquisition shifts.

In general, improvements arising from counterfactual contrastive learning are most visible in settings where learning domain-aligned representations presents a strong advantage over having domain-clustered representations (as learned in standard SimCLR). This is notably the case for scanners under-represented during finetuning, and in settings with limited label availability, where insufficient data is present to learn a `scanner-wise' decision boundary. Conversely, when a large amount of labels is available for model finetuning (e.g. up to 223k labelled samples on EMBED and 65k on PadChest), all pre-training strategies often converge to similar performance. This is explained by the fact that, with high amounts of labels, the benefit of domain-aligned representations is reduced as enough data is present at finetuning time to learn meaningful decision boundaries for each scanner, even if representations are clustered per domain. Moreover, it is important to highlight that, in the case of full model finetuning, representations may change substantially over the pre-trained representations, further explaining why downstream performance differences are often more visible in linear probing experiments. However, it is important to highlight that the counterfactual contrastive objective performs consistently either better or, depending on the number of labels, on par with existing baselines.

Naturally, gains arising from counterfactual contrastive learning are bounded by the ability to generate realistic domain changes. Generated counterfactuals must be of sufficient quality to capture the variation relevant for pre-training faithfully. Our results demonstrate that current counterfactual image generation models can already produce images of sufficient quality to significantly improve learned representation over the baseline approaches. Importantly, results in our ablation study in~\cref{sec:ablationstudy} show that even when the effectiveness of the counterfactual image generation model is moderate, CF-SimCLR remains on par with, or still outperforms the SimCLR baseline. However, as effectiveness increases, the performance on particularly under-represented scanners can substantially increase (e.g. CF-SimCLR with HVAE\texttt{+}FT on the Pristina scanner). As image synthesis models continue to improve, we may expect further improvements with counterfactual contrastive learning in future works.

Importantly, the choice of the intervention variable in the counterfactual image generation step depends on which types of shifts downstream models should be robust against. In the main experiments (\cref{sec:cfsimclr:results}), where the goal was to achieve scanner robustness, we intervened on the scanner variable. In the experiment in~\cref{sec:fair_cf}, the goal was to be robust against gender shift. Hence, we generated sex counterfactuals. A combination of multiple variables could also be considered in future work. In practice, we can summarise the process to determine which variables to intervene on, as follows:
\begin{enumerate}
    \item First, an in-depth data analysis should be performed to determine likely causes of biases and subgroup disparities, and to determine which variable the downstream representations should be robust to, i.e. should ignore.
    \item The set of identified variables in step 1, should then be discussed with domain experts to decide which variables downstream models should be robust against, and whether any interactions between these variables should be considered in the causal graph used to generate the counterfactuals.
    \item Train the counterfactual generation model with the previously identified causal graph.
    \item Use the generated counterfactuals to create cross-domain pairs in the counterfactual contrastive framework, with interventions randomly sampled among all possible values when sampling positive pairs.
\end{enumerate}

Note that our proposed approach is compatible with other generative-based augmentation methods, such as generating additional images for under-represented classes or subgroups~\citep{ktena_generative_2023,khosravi_synthetically_2024}, as we can also apply the counterfactual generation to synthetic images. Our experiments comparing SimCLR\texttt{+} with CF-SimCLR show that the proposed approach has benefits beyond complementing the training set with synthetic data and that the proposed counterfactual pair generation framework fundamentally changes the organisation of the embedding space (\cref{fig:mammo-embeddings}), as such it may complement other approaches to enhance training set diversity.

\section*{Data and code declaration}
All datasets used in this study are publicly available. PadChest~\citep{bustos_padchest_2020} can be found at \url{https://bimcv.cipf.es/bimcv-projects/padchest/}, the RSNA Pneumonia Detection Dataset~\citep{stein_rsna_2018} can be downloaded at \url{https://www.kaggle.com/c/rsna-pneumonia-detection-challenge},
CheXpert~\citep{irvin_chexpert_2019} can be found at \url{https://stanfordaimi.azurewebsites.net/datasets/8cbd9ed4-2eb9-4565-affc-111cf4f7ebe2}. We only used the publicly available part of the EMBED dataset~\citep{jeong_emory_2023} instructions can be found at  \url{https://github.com/Emory-HITI/EMBED_Open_Data/tree/main}, the 
VinDR Mammo~\citep{nguyen_vindr-mammo_2023} can be found at \url{https://vindr.ai/datasets/mammo}. All the code required to reproduce our experiments is publicly available at \url{https://github.com/biomedia-mira/counterfactual-contrastive}.

\section*{Author contributions}
M.R. and B.G. conceived and designed the experiments. M.R. and F.D.S. wrote the code. M.R. performed the experiments. All authors analysed results and wrote the paper.

\section*{Acknowledgements}
M.R. is funded through an Imperial College London President's PhD Scholarship and a Google PhD Fellowship. F.D.S., T.X. and B.G. received funding from the European Research Council (ERC) under the European Union’s Horizon 2020 research and innovation programme (grant agreement No 757173, project MIRA, ERC-2017-STG). B.G. also received support from the Royal Academy of Engineering as part of his Kheiron/RAEng Research Chair in Safe Deployment of Medical Imaging AI. We acknowledge the support of the UKRI AI programme, and the Engineering and Physical Sciences Research Council, for CHAI - EPSRC Causality in Healthcare AI Hub (grant number EP/Y028856/1). We acknowledge computational resources and support provided by the Imperial College Research Computing Service (http://doi.org/10.14469/hpc/2232).

\bibliographystyle{elsarticle-harv}
\bibliography{references_all,additional_bib}

\end{document}